\documentclass[runningheads]{llncs}

\usepackage{eccv}

\usepackage{eccvabbrv}

\usepackage{graphicx}
\usepackage{booktabs}
\usepackage{adjustbox}
\usepackage[utf8]{inputenc} 
\usepackage[T1]{fontenc}    
\usepackage{url}            
\usepackage{booktabs}       
\usepackage{amsfonts}       
\usepackage{nicefrac}       
\usepackage{microtype}      
\usepackage{xcolor}         
\usepackage{makecell}
\usepackage{dsfont}

\usepackage{algorithm}
\usepackage{algpseudocode}

\usepackage{graphicx}
\usepackage{booktabs}

\usepackage[accsupp]{axessibility}  

\usepackage{amsmath}
\usepackage{amssymb}
\usepackage{float}
\usepackage[table]{xcolor}
\usepackage[dvipsnames]{xcolor}
\usepackage{enumitem}
\usepackage{multirow}
\usepackage[mathscr]{eucal}
\usepackage{comment}
\usepackage{multicol,lipsum}
\usepackage{arydshln}
\usepackage{pifont}
\usepackage{booktabs}
\usepackage{bbding}

\usepackage{marvosym}

\def\I{{\bf I}}

\def\q{{\bf q}}
\def\s{{\bf s}}

\def\V{{\bf V}}

\def\W{{\bf W}}
\def\w{{\bf w}}
\def\0{{\bf 0}}
\def\1{{\bf 1}}

\def\RB{{\mathbb R}}
\def\NB{{\mathbb N}}

\definecolor{purple}{rgb}{0.56,0.27,0.68}
\definecolor{darkpurple}{rgb}{0.3, 0.125, 0.55} 
\definecolor{red}{rgb}{0.95,0.4,0.4}
\definecolor{purered}{rgb}{1,0,0}
\definecolor{blue}{rgb}{0.4,0.4,0.95}
\definecolor{darkblue}{rgb}{0,0,0.8}
\definecolor{grey}{rgb}{0.6,0.6,0.6}
\definecolor{col1}{RGB}{232, 161, 148}
\definecolor{col11}{RGB}{255, 228, 228}
\definecolor{col2}{RGB}{148, 187, 232}
\definecolor{col33}{RGB}{206, 239, 255}
\definecolor{col3}{RGB}{233, 255, 245}
\definecolor{lightgrey}{rgb}{0.85,0.85,0.85}
\definecolor{lightlightgrey}{rgb}{0.9,0.9,0.9}
\definecolor{verylightBG}{rgb}{0.9,0.99,0.99}
\definecolor{darkgreen}{rgb}{0., 0.85, 0.5}

\definecolor{gtred}{RGB}{204, 0, 0}
\definecolor{predgreen}{RGB}{31, 237, 31}
\definecolor{figGreen}{RGB}{56, 118, 29}

\usepackage{hyperref}

\usepackage{orcidlink}

\begin{document}

\title{Solving Semi-Supervised Few-Shot Learning from an Auto-Annotation Perspective} 

\titlerunning{SWIFT}

\author{First Author\inst{1}\orcidlink{0000-1111-2222-3333} \and
Second Author\inst{2,3}\orcidlink{1111-2222-3333-4444} \and
Third Author\inst{3}\orcidlink{2222--3333-4444-5555}}

\author{Tian Liu\inst{1} \and
Anwesha Basu\inst{1} \and
James Caverlee\inst{1} \and
Shu Kong\inst{2,3,\text{\Letter}}}

\authorrunning{T.~Liu et al.}

\institute{
$^{1}$Texas A\&M University, 
$^{2}$University of Macau, 
$^{3}$Institute of Collaborative Innovation\\
website and code: \url{https://tian1327.github.io/SWIFT} 
}

\maketitle

\begin{abstract}
Semi-supervised few-shot learning (SSFSL) resembles real-world applications such as ``auto-annotation'', as it aims to learn a model from a few labeled and abundant unlabeled task-specific examples to annotate the unlabeled ones.
Despite the availability of powerful open-source Vision-Language Models (VLMs) and open-world data, existing SSFSL literature largely neglects these resources. In contrast, the related area few-shot learning (FSL) has already exploited them to boost performance.
Arguably, to solve real-world auto-annotation, SSFSL should leverage such open resources.
To bridge this gap, we explore established SSL methods to finetune a VLM.
Unexpectedly, they significantly underperform FSL baselines that do not use unlabeled data. 
Our in-depth analysis reveals the root cause of failure:
VLMs produce ``flat'' distributions of softmax probabilities,
resulting in zero utilization of unlabeled data and weak supervision signals.
To address this challenge, we propose an embarrassingly simple solution that uses temperatures to sharpen the softmax output, which not only increases the confidence scores of pseudo-labels to improve the utilization of unlabeled data, but also strengthens training supervision for effective finetuning.
Furthermore, we exploit task-relevant open data, e.g., those retrieved from VLMs' publicly available pretraining set. To mitigate the imbalance and domain gaps in retrieved data, we employ a stage-wise training strategy.
Building on the successful finetuning of VLMs and the exploitation of open data, we present a simple yet effective SSFSL method, \textbf{S}tage-\textbf{Wi}se \textbf{F}inetuning with \textbf{T}emperatures (\textbf{SWIFT}).
Across five benchmarks, SWIFT outperforms recent FSL and SSL methods by $\sim$5 accuracy points. SWIFT even rivals supervised learning, which finetunes a VLM assuming unlabeled data having ground-truth labels!

\end{abstract}

\section{Introduction}
\label{sec:introduction}

\begin{figure}[t]
\centering
\includegraphics[width=0.95\linewidth, clip=true,trim = 0mm 0mm 0mm 0.5mm]
{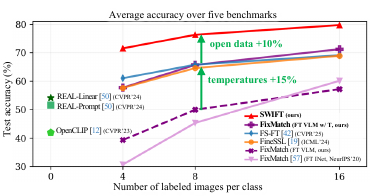}
\vspace{-1mm}
\caption{\small 
\textbf{
Benchmarking SSFSL methods across five datasets.
}
First, traditional \textcolor{purple}{SSL methods} e.g., FixMatch \cite{sohn2020fixmatch}, which finetune ImageNet-pretrained backbones \cite{he2016deep},  
underperform recent VLM-based \textcolor{Green}{zero-shot methods} like REAL~\cite{parashar2024neglected}.
Second, \textcolor{orange}{SOTA SSL method FineSSL}  \cite{gan2024erasing} exploits a frozen VLM via prompt learning, achieving notable gains yet underperforming recent \textcolor{darkblue}{few-shot method FS-FT} \cite{liu2025few}, which finetunes a VLM without using unlabeled data.
Third, we are thus motivated to \textcolor{darkpurple}{finetune a VLM using established FixMatch} (denoted ``FT VLM'') and find that it fails, due to the ``flat'' softmax probabilities of VLMs (\cref{fig:diagnosis}).
Fourth, to address this challenge, we \textcolor{darkpurple}{adopt simple temperatures to finetune VLM} (denoted ``FT VLM w/ T''), which significantly improves accuracy by >15\%, outperforming FineSSL \cite{gan2024erasing}.
Finally, \textcolor{purered}{our method SWIFT} further exploits open data via stage-wise training, achieving further 10\% accuracy gains.
}
\label{fig:benchmark}
\vspace{-3mm}
\end{figure}

Semi-Supervised Few-Shot Learning (SSFSL) \cite{dong2024pseudo, li2019learning, ling2022semi, wei2022embarrassingly} is well-suited for real-world applications such as ``auto-annotation'' \cite{liu2025few, madan2024revisiting, ma2025towards}.
It aims to learn a model utilizing a small set of task-specific labeled data and a much larger set of unlabeled data to produce reliable annotations for the unlabeled examples.

\begin{figure}[t]
  \centering
  \begin{minipage}[c]{0.54\linewidth}
    \centering
    \includegraphics[width=0.997\linewidth, clip=true, trim=0mm 0mm 0mm 0mm]{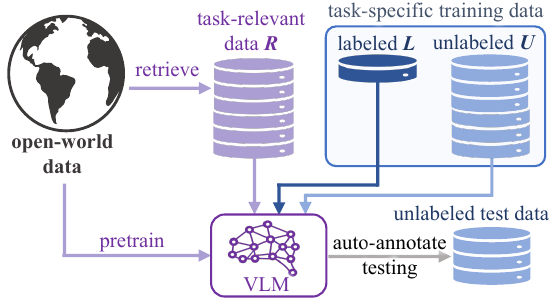}
  \end{minipage}
  \hfill 
  %
  \begin{minipage}[c]{0.42\linewidth}
    \small
    \caption{
      \textbf{A realistic setup for SSFSL} 
      embraces open-source VLMs and open-world data.
      It allows retrieving task-relevant data \textbf{$R$} to augment the task-specific labeled and unlabeled training data \textbf{$L$} and \textbf{$U$}.
      Applying the learned model to test data amounts to using it to auto-annotate the unlabeled test data.
    }
    \label{fig:problem-setup}
  \end{minipage}
  \vspace{-4mm}
\end{figure}

\textbf{Status Quo.}
SSFSL extends few-shot learning (FSL) by leveraging unlabeled examples for training, alongside a small labeled set.
Modern FSL methods \cite{liu2025few, clap24, wang2025robust} have achieved significant progress by finetuning pretrained Vision-Language Models (VLMs) and retrieving task-relevant open data (e.g., from VLM's publicly available pretraining set) to augment training data.
In contrast, contemporary SSFSL and general Semi-Supervised Learning (SSL) methods largely overlook these powerful resources.
Some of them either train models from scratch~\cite{zhang2021flexmatch, sohn2020fixmatch, Berthelot2020ReMixMatch, berthelot2019mixmatch} or finetune ImageNet-pretrained backbones~\cite{su2021realistic, wang2022usb, chen2023softmatch, wang2022debiased}.
Other works \cite{dong2024pseudo, ling2022semi, wei2022embarrassingly, li2019learning} adopt a simulated setup, first pretraining a ResNet-12 model on a large ``base'' dataset and then semi-supervised finetuning on domain-specific data.
A few recent SSL methods exploit a frozen VLM with prompt learning rather than finetuning \cite{zhang2025revisiting, gan2024erasing, zhang2024candidate, menghini2023enhancing}.
These design choices significantly limit the progress of SSFSL.
To solve real-world applications like auto-annotation, 
we study the underexplored paradigm that explores finetuning VLMs and retrieving open data for SSFSL (\cref{fig:problem-setup}).

\textbf{Motivation.}
SSFSL frames the important application ``auto-annotation'' as both aim to exploit limited labeled data and abundant unlabeled data to produce labels for unlabeled examples.
Auto-annotation also inspires recent FSL research, leading to a rigorous and practical paradigm that prioritizes prediction accuracy over parameter efficiency~\cite{liu2025few, madan2024revisiting} and eschews the unrealistic assumption of accessible large validation sets
for hyperparameter tuning~\cite{lin2023multi, clap24, liu2025few}.
Moreover, concurrent advances in zero-shot learning (ZSL)~\cite{liu2023learning, wallingford2023neural, iscen2023retrieval,  saha2024improved, parashar2024neglected} and FSL~\cite{liu2025few, wang2025robust} 
have demonstrated that exploiting not only VLMs but also open data (e.g., VLM's pretraining data) can substantially improve performance.
To bridge the research gap,
we rigorously explore SSFSL by finetuning open-source VLMs and 
exploiting open data (\cref{fig:problem-setup}).

\textbf{Challenges and Insights.}
Inspired by the significant improvement in FSL by finetuning a VLM on few-shot labeled data \cite{liu2025few}, we explore established SSL methods to finetune a VLM by further utilizing unlabeled data.
Unexpectedly, such direct adoption underperforms FSL and even ZSL baselines (\cref{fig:benchmark}),
aligning with observations in prior work~\cite{zhang2025revisiting}.
We identify the root cause: \emph{VLMs produce ``flat'' distribution of softmax probabilities} (\cref{fig:diagnosis}A), which results in zero utilization of unlabeled data (\cref{fig:diagnosis}B) and weak learning signals that prevent effective finetuning (\cref{fig:diagnosis}C). 
To address this, we propose an embarrassingly simple technique that uses temperatures to sharpen softmax probabilities, significantly improving SSFSL performance (\cref{tab:compare_sota}).
Moreover, we retrieve task-relevant open data and employ stage-wise training to mitigate the inherent noise, imbalance, and out-of-distribution (OOD) nature of retrieved data.
Integrating these insights, our method, \textbf{S}tage-\textbf{Wi}se \textbf{F}inetuning with \textbf{T}emperatures ({\bf SWIFT}),
achieves state-of-the-art (SOTA) performance across five SSFSL benchmarks (\cref{fig:benchmark}).

\textbf{Contributions}.
We make three key contributions. 
\begin{itemize}[label=$\bullet$,  topsep=0pt]

\item From an auto-annotation perspective, we explore an underexplored 
SSFSL paradigm by exploiting open-source models and open data. Our realistic setup (\cref{fig:problem-setup}) significantly advances SSFSL performance.

\item We identify why existing SSL methods fail to finetune VLMs: VLMs produce ``flat'', low confidence scores that lead to zero utilization of unlabeled data and weak supervision.
We address this by leveraging temperatures.

\item We present a simple yet effective method, \textbf{SWIFT}, 
which outperforms prior FSL and SSL approaches by $>$5 accuracy points across five SSFSL datasets, even rivaling fully supervised learning.

\end{itemize}

\section{Related Works}
\label{sec:related_works}

\textbf{Semi-Supervised Few-Shot Learning} (SSFSL) \cite{dong2024pseudo,li2019learning, ling2022semi,wei2022embarrassingly}
is a special case of SSL~\cite{chapelle2009semi} in which the amount of unlabeled data is extremely limited.
Classic SSL methods build on several core ideas:
consistency regularization~\cite{yang2025unimatch, wei2023towards, zheng2022simmatch}, which enforces prediction consistency between strong and weak augmentations of unlabeled examples;
pseudo-labeling \cite{cascante2021curriculum, arazo2020pseudo, lee2013pseudolabel, xie2020self}, which uses a teacher model's predictions to supervise the student model;
and transfer learning \cite{chen2020big, he2020momentum}, where a model is first self-supervised, pretrained on large unlabeled datasets, and then adapted to a downstream task with labeled data.
Among various SSL methods, FixMatch \cite{sohn2020fixmatch} stands out for its simplicity and strong performance.
Subsequent works improve FixMatch by selecting confident pseudo-labels with adaptive thresholds \cite{zhang2021flexmatch, xu2021dash, huang2021pseudo, guo2022class, berthelot2022adamatch, wang2023freematch, chen2023softmatch, du2024simpro}, or enhancing pseudo-label quality through logit adjustment \cite{wang2022debiased, menon2021long} or teacher ensembling \cite{cai2021exponential, laine2017temporal, tarvainen2017mean}.
Recent SSL methods further leverage zero-shot predictions from pretrained VLMs as auxiliary supervisions \cite{chung2025enhancing, wang2022debiased}, or adopt prompt tuning with a frozen VLM \cite{zhang2025revisiting, gan2024erasing, zhang2024candidate, menghini2023enhancing}.
However, no existing works have successfully finetuned VLMs for SSL. 
The recent work
\cite{zhang2025revisiting} even asserts that tailored SSL methods are required to finetune VLMs effectively.
Our study, however, reveals the root cause of such failures is the ``flat'' softmax outputs from VLMs. We address this by leveraging temperatures in learning.
For the first time, we enable successful finetuning of VLM for SSFSL.

\textbf{Open-World Data and Open-Source Models.}
The abundance of open-world data not only facilitates training more robust and generalizable models~\cite{kong2021opengan, hendrycks2018deep, shen2025solving}, but also enables pretraining foundational models~\cite{radford2021learning, align, xu2023metaclip, cherti2023reproducible, sun2024alpha}.
By leveraging pretrained VLMs,
recent methods in ZSL \cite{parashar2024neglected, wallingford2023neural, liu2023learning} and FSL \cite{clap24, lin2023multi, wang2025robust, clipadapter, zhang2025enhancing} have achieved substantial improvements.
These paradigms \cite{liu2025few, liu2023learning, wallingford2023neural, iscen2023retrieval,  saha2024improved} further enhance performance by retrieving task-relevant open data, such as those from VLMs' pretraining datasets \cite{laion400m, laion5b}. 
In contrast, existing SSL \cite{berthelot2022adamatch,sohn2020fixmatch,wang2022debiased,zhang2021flexmatch}  and SSFSL \cite{ dong2024pseudo, ling2022semi, ma2025tenet, wei2022embarrassingly} research largely overlooks these powerful resources. Current frameworks typically rely on training from scratch, finetuning ImageNet-pretrained backbones, or employing frozen VLMs without finetuning \cite{zhang2025revisiting, gan2024erasing, zhang2024candidate, menghini2023enhancing}.
Intuitively, combining VLM finetuning with open-world data retrieval would significantly enhance SSFSL. However, this potential remains largely untapped. 
We propose an SSFSL method that integrates both VLM finetuning and open data retrieval,
thereby substantially advancing SSFSL research.

\textbf{Temperature}
is a hyperparameter used in the softmax function to control the sharpness of the resulting class probability distributions \cite{hinton2015distilling, jang2016categorical}.
It has been applied in different tasks for various purposes,
such as softening the teacher model's logits to enhance knowledge distillation~\cite{hinton2015distilling}, 
calibrating model confidence \cite{guo2017calibration, tu2023closer, tu2024empirical},
adjusting sampling diversity in language generation \cite{holtzman2020curious}, 
and scaling contrastive loss to facilitate large-scale pretraining of foundation models \cite{chen2020simple, he2020momentum, wang2017normface, wu2018unsupervised}.
In VLMs' pretraining  \cite{cherti2023reproducible, radford2021learning, xu2023metaclip}, 
temperature is set as a learnable parameter in the contrastive loss, 
initialized to 0.07~\cite{wu2018unsupervised} and eventually clipped at 0.01 to stabilize training \cite{radford2021learning}.
With pretrained VLMs, 
recent methods commonly inherit the fixed temperature of 0.01 
when initializing classifier weights in ZSL \cite{wortsman2022robust} or when computing cross-entropy loss in FSL for learning the classifier or prompt \cite{clap24, zhang2024candidate, lin2023multi, tipadapter, clipadapter, zhou2022learning}.
The significance of temperature in VLM finetuning remains unexplored.
Our work, for the first time, demonstrates that temperature is crucial for successful SSFSL with VLMs.

\section{Problem Formulation and Methods}
\label{sec:problem_formulation_and_methods}
In this section, we first formalize our SSFSL setup. 
Then, we repurpose representative SSL and FSL methods as baselines to finetune a VLM for SSFSL and analyze their failures. 
Lastly, we present a simple temperature-based solution and derive our final SSFSL method.

\textbf{Problem Setup and Notations.}
A downstream $C$-way image classification task provides a small set of labeled images $L=\{(\I_i, y_i)\}_{i=1}^{N_l}$, where $y_i \in \NB^C$, and a large set of unlabeled images $U=\{\I_j\}_{j=1}^{N_u}$ with $N_l \ll N_u$.
Each class in $L$ contains $K \in \{4,8, 16\}$ labeled images, formalizing a $K$-shot setting with $N_l=K*C$.
A pretrained VLM and an open data pool (e.g., VLM's pretraining dataset) $P$ are provided.
The VLM consists of a visual encoder $\mathbf{V}(\cdot)$ and a text encoder $\mathbf{T}(\cdot)$,
which transform an input image $\I$ into visual embedding $\V(\I) \in \RB^{d} $ and its class name into text embedding $\mathbf{T}(y)\in \RB^{d}$, respectively.
From $P$, one can retrieve a set of task-relevant examples $R=\{(\I_i, y_i)\}_{i=1}^{N_r}$.
These examples naturally follow imbalanced distributions and contain noisy labels 
$y_i\in \NB^C$, exhibiting distribution shifts relative to the 
task-specific ones in $L$ and $U$ \cite{liu2025few}.\footnote{As we focus on VLM finetuning rather than data retrieval, we adopt the method proposed in~\cite{parashar2024neglected, liu2025few} to retrieve task-relevant data.}
Our SSFSL exploits data in $L$, $U$, and $R$ to finetune the VLM and evaluate on the held-out test set (Fig.~\ref{fig:problem-setup}).

\textbf{Validation-free Evaluation Protocol.}
Given the scarcity of labeled data, our SSFSL setup adopts a realistic \emph{``validation-free protocol''} that eschews a validation set for hyperparameter tuning. Our approach is consistent with recent works in FSL \cite{liu2025few, clap24} and SSL \cite{zhang2025revisiting, su2021realistic} that emphasize real-world applicability.
Specifically, these works employ a rigorous ``cross-dataset tuning'' strategy, in which hyperparameters (e.g., learning rate, weight decay) are tuned on a single source dataset and then directly applied to all other datasets.
Prior work \cite{liu2025few} shows that such protocol yields robust performance, without suffering from overfitting even in the low-data regime.
Our experiment follows this rigorous protocol by directly adopting hyperparameters from \cite{liu2025few} and applying them throughout our experiments on all datasets, without any tuning.

\subsection{Baseline Results and Failure Diagnosis}
\label{ssec:baseline_methods}

We experiment with representative FSL and SSL methods as baselines to finetune VLM for SSFSL,
aiming to diagnose their failures and derive our method.
Results presented below are averaged over five benchmarks using the open-source VLM OpenCLIP ViT-B/32, as detailed in \cref{sec:experiments_and_results}.

\textbf{FSL Method.}
Few-shot finetuning (FS-FT) the VLM's visual encoder $\V(\cdot)$ is a competitive validation-free FSL method~\cite{liu2025few}.
First, it initializes a classifier $\W=[\w^1,  \dots, \w^C] \in \RB^{d \times C}$ using the text embeddings of the $C$ class names.
For an input image $\I$, the visual encoder encodes the image and passes it to the classifier, producing a logit vector $\q=\W^T \V(\I) \in \RB^{C\times 1}$.
A temperature parameter $\tau$ is then applied to the logits to compute the softmax probability vector $\s$, where $s^{c} =  \frac{\exp\left(q^{c} / \tau \right)}{\sum_{j=1}^{C} \exp\left(q^{j} / \tau \right)}$. Finally, $\V(\cdot)$ and $\W$ are jointly finetuned by minimizing the cross-entropy (CE) loss over the few-shot labeled set $L$: $\mathcal{L}_l = \frac{1}{\vert L\vert} \sum_{(\I,y)\in L} \ell (\W^T\V(\I)/\tau, y)$.

\textbf{SSL Methods.} 
The influential FixMatch \cite{sohn2020fixmatch} generates pseudo-labels for unlabeled data and selects high-confident ones for training (\cref{fig:temperature-learning-with-FixMatch}).
For each unlabeled image $\I \in U$, FixMatch generates a weakly augmented view $\I^w$ and passes it through the training model to produce a 
$C$-dim logits $\q^w=\W^T\V(\I^w)$.
Based on the logits $\q^w$, the softmax probabilities are calculated 
with a default temperature of 1.0: $\s^w=\text{softmax}(\q^w)$. Then, 
the pseudo-label $\hat y=\arg\max_c(\s^w)$ is derived.
Notably, to mitigate noisy pseudo-labels, FixMatch only utilizes pseudo-labels with softmax confidence higher than a threshold, i.e., $\max_c(\s^w) \ge \sigma$ (default $\sigma=0.8$ \cite{su2021realistic}).
Based on the selected pseudo-labeled data, FixMatch finetunes the model on a strongly augmented view $\I^s$ by minimizing  $\mathcal{L}^s =
\frac{1}{\vert U \vert}
\sum_{\I \in U} 
\mathds{1}[\max_c(\s^w) \ge \sigma ] \cdot \ell(\W^T\V(\I^s), \hat y)$.
$\mathcal{L}_l$ and $\mathcal{L}^s$ are jointly used in learning.
DebiasPL \cite{wang2022debiased} extends FixMatch by adjusting the logits of unlabeled data and incorporating adaptive class margins to mitigate the imbalanced distribution of pseudo-labels.
Importantly, both FixMatch \cite{sohn2020fixmatch} and DebiasPL \cite{wang2022debiased} typically finetune an ImageNet-pretrained backbone.
We conduct experiments and analyze their performance when finetuning a VLM, and compare them with
the FSL method FS-FT \cite{liu2025few} introduced above.

\begin{table}[t]
\centering
\small
\caption{\small 
\textbf{
Established SSL methods fail to finetune VLM. 
}
While directly running FixMatch \cite{sohn2020fixmatch} and DebiasPL \cite{wang2022debiased} to finetune the VLM OpenCLIP ViT-B/32 \cite{cherti2023reproducible} improves performance with the ImageNet-pretrained ResNet-50 backbone \cite{su2021realistic} under 4- and 8-shot settings,
doing so yields lower accuracy in the 16-shot setting.
Notably, it significantly underperforms FS-FT, which does not use unlabeled data for training.
\textcolor{Red}{Red superscripts} denotes accuracy degradation w.r.t. FS-FT \cite{liu2025few}.
}
\vspace{-3mm}
\label{tab:fixmatch_direct}
\setlength{\tabcolsep}{2.3mm}
\scalebox{0.85}{
\begin{tabular}{lcllll}
\toprule
\multirow{2}{*}{method} & \multirow{2}{*}{training data} & \multirow{2}{*}{backbone} & \multicolumn{3}{c}{mean acc. over five datasets} \\
\cmidrule{4-6}
& &  &4-shot &8-shot &16-shot \\
\midrule
\rowcolor{col33}
FS-FT \cite{liu2025few} & $L$ & VLM-OpenCLIP & 61.0 & 65.8 & 69.1 \\
\midrule

FixMatch \cite{sohn2020fixmatch} & $L+U$  & INet-pretrained &30.7  &45.3  &60.1  \\

FixMatch & $L+U$ & VLM-OpenCLIP
& 39.3$^{\textcolor{Red}{-21.7}}$ 
& 49.9$^{\textcolor{Red}{-15.9}}$ 
& 57.2$^{\textcolor{Red}{-11.9}}$ \\

\midrule

DebiasPL \cite{wang2022debiased} & $L+U$  & INet-pretrained
&34.5  &49.0  &62.7  \\

DebiasPL & $L+U$
& VLM-OpenCLIP
&39.6 $^{\textcolor{Red}{-21.4}}$ 
&49.8 $^{\textcolor{Red}{-16.0}}$ 
&57.1 $^{\textcolor{Red}{-12.0}}$ \\

\bottomrule
\end{tabular}}
\vspace{-4mm}
\end{table}

\begin{figure}[t]
\centering
\includegraphics[width=0.99 \textwidth, clip=true, trim = 0mm 7mm 4mm 7mm]{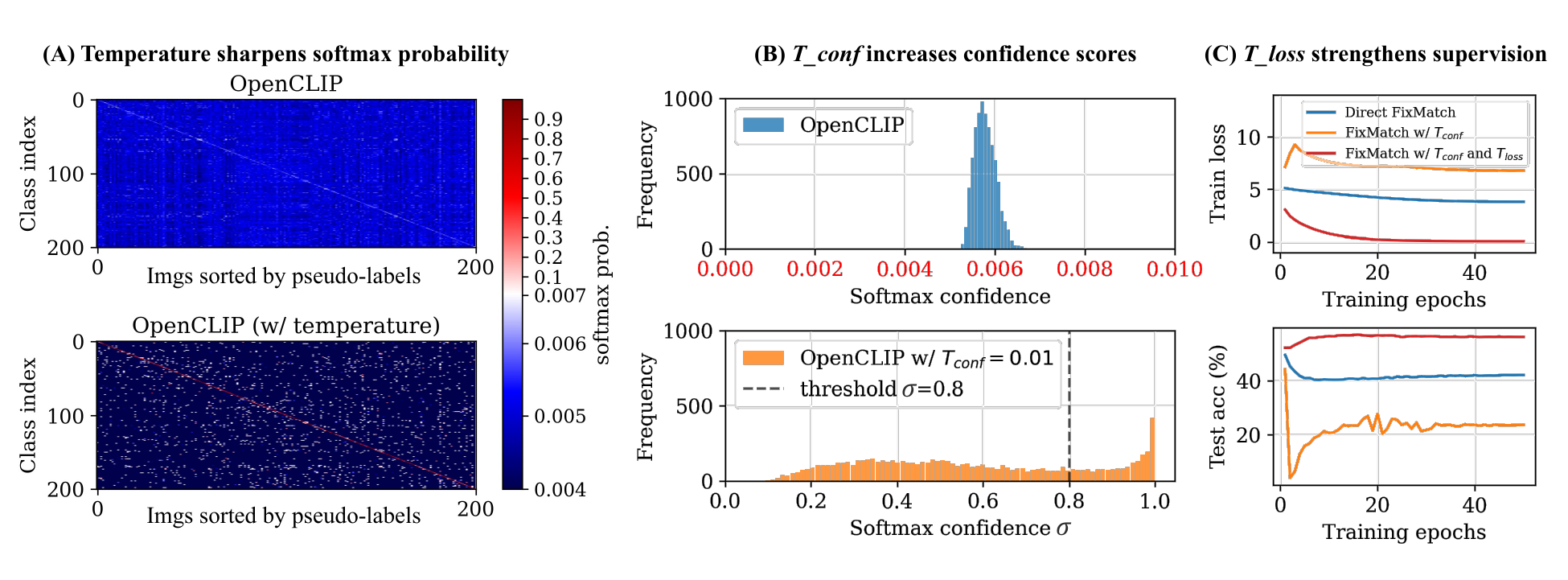}
\vspace{-3mm}
\caption{\small 
{\bf Challenges in SSFSL with VLMs and our solution via temperatures.}
\textbf{(A-top)} We obtain softmax probabilities of the unlabeled images from semi-Aves dataset
by zero-shot prompting the OpenCLIP ViT-B/32 model with its 200 class names.
VLMs produce a ``flat'' distribution of softmax probabilities (200-dim column vector per unlabeled example), resulting in low confidence scores \textbf{(B-top)} and
weak training supervision that prevents effective finetuning \textbf{(C)}.
We solve this by using simple temperatures to sharpen the softmax distributions, as illustrated by the red diagonal line in \textbf{(A)}-bottom.
Concretely, we apply a temperature $T_\text{conf}$ (e.g., 0.01) to increase pseudo-label confidence, enabling the utilization of unlabeled data using the default threshold of 0.8 \textbf{(B-bottom)}. 
We also use a temperature $T_\text{loss}$ to compute CE loss, enhancing training loss reduction and improving test accuracy (red lines in \textbf{(C)}).
}
\label{fig:diagnosis}
\vspace{-4mm}
\end{figure}

\textbf{Failure Diagnosis.}
Counterintuitively, our experiments in \cref{tab:fixmatch_direct} show that finetuning VLM using FixMatch and DebiasPL significantly underperforms the simple FSL method that does not exploit unlabeled data. Our results align with the observation in recent work \cite{zhang2025revisiting}, although the underlying reason remains uninvestigated.
Remarkably, our in-depth analysis identifies the root cause: \emph{VLMs produce ``flat'' distributions of softmax probabilities} (\cref{fig:diagnosis}A), which:
\begin{enumerate}[topsep=0pt]
    \item 
    yields low-confidence pseudo-labels, resulting in zero utilization of unlabeled data with a default high confidence threshold (Fig.~\ref{fig:diagnosis}B);

    \item 
    provides weak training supervision, preventing effective finetuning (\cref{fig:diagnosis}C).
    
\end{enumerate}

\subsection{Temperature is the Remedy}
\label{sec:remedies}

To address the above issues,
we present an embarrassingly simple technique by using temperatures, which have not been explored in the SSFSL literature.\footnote{It is worth noting that the recent work \cite{zhang2025revisiting} uses FixMatch to finetune the VLM CLIP, achieving worse performance than finetuning on labeled data only.
Importantly, their released code \cite{OSU_MLB_SSLFoundationModels_vit_no_temperature} shows that they do not use temperatures.
}

{\bf Confidence Temperature Enables Utilization of Unlabeled Data.} 
To address the small confidence scores from VLM, we use a confidence temperature $T_{\text{conf}}$ to sharpen the softmax probability distributions for the weakly augmented images $\I^w$:
$\s^w =\text{softmax}(\W^T\V(\I^w)/T_{\text{conf}})$.
Lowering the temperature increases pseudo-label confidence, enabling the utilization of unlabeled data with the default high-confidence threshold (\cref{fig:diagnosis}B-bottom).
One may argue that an alternative approach is to lower the confidence threshold $\sigma$.
However, setting an appropriate threshold requires careful tuning on a validation set \cite{zhang2025revisiting}, which is impractical under the realistic ``validation-free'' setup and nearly infeasible given the narrow range of confidence scores (\cref{fig:diagnosis}B-top). 
In contrast, our analysis demonstrates that using a confidence temperature is much more robust, where setting $T_{\text{conf}}$ within a broad range [0.001, 0.05] consistently yields substantial performance gains (\cref{fig:confidence_temp_fixmatch}).

{\bf Loss Temperature Strengthens Training Supervision.} 
While $T_\text{conf}$ enables utilization of unlabeled data, applying it alone yields much higher training loss and lower test accuracy (\cref{fig:diagnosis}C), due to the weak training supervision.
To strengthen the supervision from selected pseudo-labeled data, we apply a loss temperature $T_{\text{loss}}$ when computing the CE loss between pseudo-labels and strongly augmented images $\I^s$:
$\mathcal{L}_u = 
\frac{1}{\vert U \vert} 
\sum_{\I \in U} 
\mathds{1}[\max_c(s^c) \ge \sigma ] \cdot \ell(\W^T\V(\I^s) / T_{\text{loss}}, \hat y)$.
Similarly, we apply $T_{\text{loss}}$ when calculating the CE loss for labeled data: $\mathcal{L}_l = \frac{1}{\vert L \vert} \sum_{(\I, y) \in L} \ell(\W^T\V(\I) / T_{\text{loss}}, y)$. 
Crucially, rather than treating $T_\text{loss}$ as a static hyperparameter, we demonstrate that learning $T_\text{loss}$ dynamically during training further enhances SSFSL performance (\cref{fig:loss_temp_fixmatch}). 

\begin{figure}[t]
  \centering
  \small
   \includegraphics[width=0.99\linewidth, clip=true, trim = 0mm 0mm 0mm 0mm]{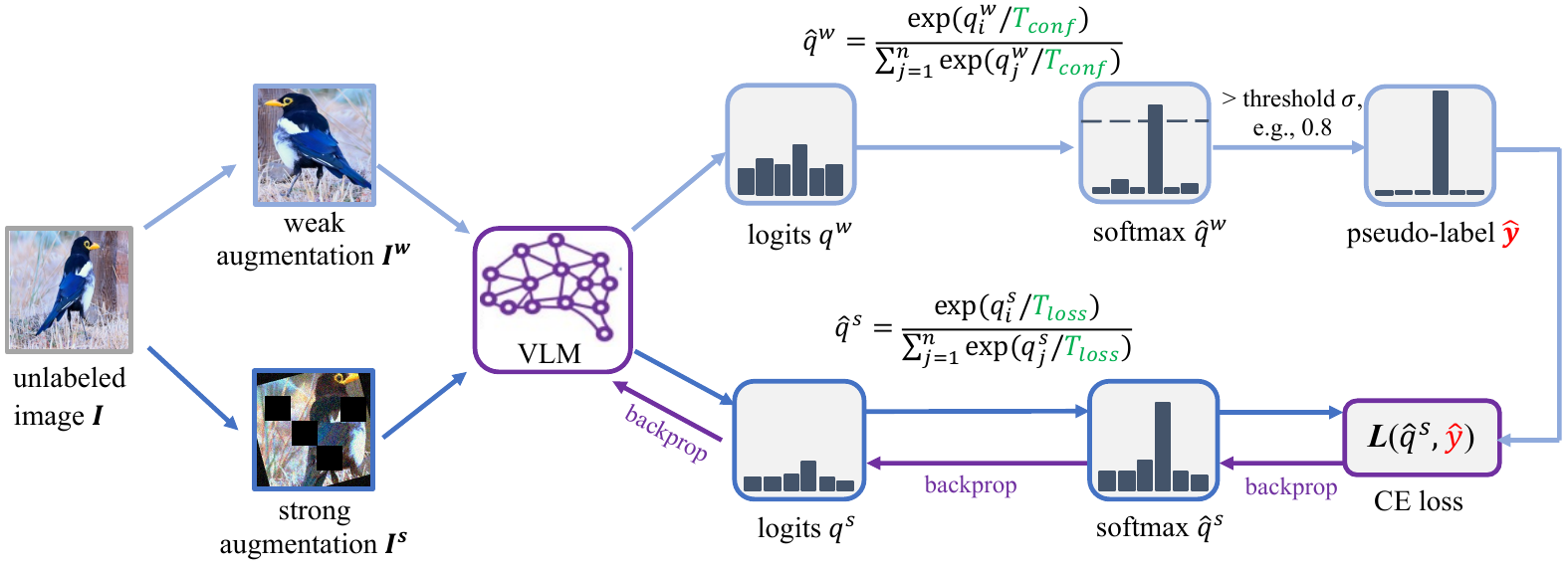}
  \vspace{-2mm}
  \caption{\small
  {\bf Illustration of our temperature-based solution for finetuning a VLM with FixMatch.}
  First, $T_{\text{conf}}$ sharpens softmax probabilities and increases confidence scores of pseudo-labels (\cref{fig:diagnosis}B), increasing the utilization of unlabeled data.
  Second, $T_{\text{loss}}$ strengthens the supervision from CE loss, enabling effective VLM finetuning.  
  We provide PyTorch-style pseudo-code in the Supplement. 
  }
  \vspace{-4mm}
  \label{fig:temperature-learning-with-FixMatch}
\end{figure}

\cref{fig:temperature-learning-with-FixMatch} illustrates our temperature-based solution applied to FixMatch. \cref{fig:diagnosis}C confirms that applying $T_{\text{conf}}$ and
$T_{\text{loss}}$ accelerates learning and improves test accuracy. 
Notably, our simple solution is compatible with many subsequent FixMatch-based SSL methods, e.g., DebiasPL \cite{wang2022debiased} (\cref{tab:compare_sota}).
Post-analyses in Figures \ref{fig:loss_temp_fixmatch} and \ref{fig:confidence_temp_fixmatch}  demonstrate SWIFT's robustness to temperature settings.

\subsection{Exploiting Open Data via SWIFT}

Building on our successful finetuning of VLMs, we further exploit open data to enhance SSFSL.
Retrieval augmentation (RA) has been studied in ZSL \cite{liu2023learning, wallingford2023neural, iscen2023retrieval,  saha2024improved} and FSL \cite{liu2025few} literature, but remains unexplored in SSL.
Specifically, prior ZSL and FSL works retrieve task-relevant data from VLMs' publicly available pretraining sets, e.g., LAION \cite{laion400m, laion5b}.
For fair comparisons with prior methods and to ensure reproducibility, we follow \cite{parashar2024neglected, liu2025few} to retrieve from LAION-400M \cite{laion400m} and construct a set of noisy labeled data $R$ (\cref{fig:problem-setup}).

\begin{figure}[t]
  \centering
  \small
  \includegraphics[width=0.99\linewidth, clip=true, trim = 0mm 0mm 0mm 0mm]{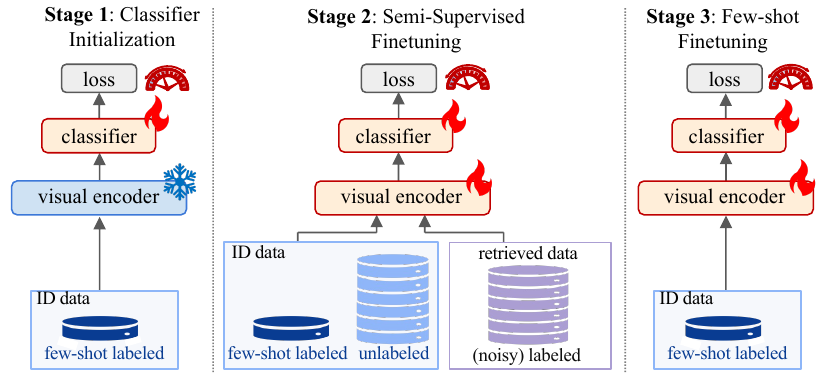}
  \caption{\small
  {\bf Illustration of our SWIFT training pipeline for SSFSL}.
  SWIFT adopts a stage-wise training approach, applying temperature in all three stages to enable effective VLM finetuning. SWIFT exploits not only task-specific few-shot labeled data $L$ and abundant unlabeled data $U$ but also large amounts of retrieved open data $R$.
  }
  \vspace{-4mm}
  \label{fig:swift}
\end{figure}

With $R$,
we employ a CE loss: $\mathcal{L}_r = \frac{1}{\vert R\vert} \sum_{(\I,y)\in R} \ell (\W^T\V(\I)/ T_{\text{loss}}, y)$, in addition to the losses $\mathcal{L}_l$ and $\mathcal{L}_u$.
A straightforward approach is to finetune the VLM using all three losses $\mathcal{L}_l$, $\mathcal{L}_u$, and $\mathcal{L}_r$ in a single stage.
However, such single-stage training is suboptimal because the retrieved data $R$ exhibit imbalanced class distributions, noisy labels, and distribution shifts relative to task-specific images $L$ and $U$ \cite{liu2025few} (see examples in Supplement \cref{fig:example_data}).
To address these issues, we propose a stage-wise SSFSL method, \textbf{SWIFT}: \textbf{S}tage-\textbf{Wi}se \textbf{F}inetuning with \textbf{T}emperatures (\cref{fig:swift}), consisting of three training stages, \emph{where our employed temperatures consistently yields remarkable improvements}:

\begin{itemize}[label=$\bullet$, topsep=0pt, partopsep=0pt] 
\item {\em Stage-1 Classifier Initialization.}
While existing FSL methods \cite{liu2025few, clap24, lin2023multi} typically initialize the classifier weights $\W$ using text embeddings of class names, we find that further learning the classifier on few-shot labeled data $L$ with a loss temperature $T_\text{loss}$ notably improves subsequent finetuning (\cref{tab:classifier_init}).

\item {\em Stage-2 Semi-Supervised Finetuning.}
We finetune both visual encoder $\V(\cdot)$ and classifier $\W$ using a SSL method (e.g., FixMatch \cite{sohn2020fixmatch}) combining three losses $\mathcal{L}_l$, $\mathcal{L}_u$, and $\mathcal{L}_r$.
Crucially, we use a confidence temperature $T_\text{conf}$ and a loss temperature $T_\text{loss}$ to ensure successful SSL (\cref{fig:temperature-learning-with-FixMatch}).

\item {\em Stage-3 Few-shot Finetuning.}
We finetune  $\V(\cdot)$ and classifier $\W$ on few-shot labeled data $L$ again, with a loss temperature $T_\text{loss}$.
Doing so effectively mitigates domain gaps, class imbalance, and noisy labels in $R$ \cite{liu2025few}.

\end{itemize}

\section{Experiments and Results}
\label{sec:experiments_and_results}

We validate SWIFT by benchmarking it against recent FSL and SSL methods.
We also conduct post-analyses to demonstrate the robustness of SWIFT to temperature settings.

\subsection{Experimental Setup}
\label{sec:exp_setup}

{\bf Datasets and Metrics.}
Following the auto-annotation perspective~\cite{liu2025few} and recent SSL studies \cite{zhang2025revisiting}, we evaluate on challenging fine-grained recognition datasets where pretrained VLMs, such as OpenCLIP \cite{radford2021learning}, yield low zero-shot accuracy.
This also suggests that their pretraining data are less aligned with these datasets, thereby reducing the risk of data leakage when adopting retrieval augmentation (RA).
The datasets include semi-Aves \cite{semi-aves} (CC-BY 4.0 License), FGVC-Aircraft \cite{aircraft}, Stanford Cars \cite{cars} (Custom Non-commercial License), EuroSAT \cite{eurosat} (MIT License), and DTD \cite{dtd} (Custom Research-Only License). 
These datasets span bird species, aircraft and car models, land-use and cover in satellite imagery, and texture types (details and examples in Supplementary \cref{tab:datasets} and \cref{fig:example_data}). 
For each method, we report its mean accuracy across five benchmark test sets in the main paper and provide detailed results in the Supplement.
For RA, we retrieve data for each dataset from the publicly available LAION-400M \cite{laion400m}.

{\setlength{\tabcolsep}{0.2em}
\begin{table}[t]
\begin{minipage}{0.6\textwidth}
    \centering
    \vspace{-3mm}
    \scalebox{0.77}{
\begin{tabular}{clllll}
\toprule
 &\multirow{2}{*}{method}  & \multirow{2}{*}{\textcolor{Gray}{\makecell{}}} & \multicolumn{3}{c}{mean acc. over five datasets} \\
 \cmidrule{4-6}
  & &   & 4-shot & 8-shot & 16-shot \\

\midrule

\multirow{10}{*}{\rotatebox[origin=c]{90}{\makecell{FSL}}} & CoOp \cite{zhou2022learning} \textsubscript{IJCV'22} & \multirow{1}{*}{\makecell{\textcolor{Gray}{\makecell{PL}}}}  & 48.8 & 54.6 & 59.5 \\

 & PLOT \cite{chen2022plot} \textsubscript{ICLR'23}  &\textcolor{Gray}{PL}  & 50.3 & 54.9 & 59.3 \\
 \cmidrule{2-6}

 & Linear Probing \cite{radford2021learning} \textsubscript{ICML'21}  & \multirow{1}{*}{\textcolor{Gray}{\makecell{AL}}}  &57.0 &61.2 &64.7 \\
 & CLIP-Adapter \cite{clipadapter} \textsubscript{IJCV'23} &\textcolor{Gray}{AL}  & 48.6 & 55.2 & 60.0 \\
 & Tip-Adapter (f) \cite{tipadapter} \textsubscript{ECCV'22}  &\textcolor{Gray}{AL}  & 50.0 & 52.9 & 58.0 \\
 & TaskRes(e) \cite{yu2023task} \textsubscript{ECCV'22}  &\textcolor{Gray}{AL}  & 53.0 & 58.0 & 62.0 \\
 & CMLP \cite{lin2023multi} \textsubscript{CVPR'23} &\textcolor{Gray}{AL}  & 54.4 & 59.4 & 64.2 \\
 & CLAP \cite{clap24} \textsubscript{CVPR'24}  &\textcolor{Gray}{AL}  & 56.9 & 61.3 & 65.5 \\
 \cmidrule{2-6}

& FS-FT \cite{liu2025few} \textsubscript{CVPR'25} & \multirow{1}{*}{\textcolor{Gray}{\makecell{FT}}}  & 61.0 & 65.8 & 69.1 \\

&\cellcolor{col33}SWAT  \cite{liu2025few} \textsubscript{CVPR'25} (w/ RA) 
&\cellcolor{col33}\textcolor{Gray}{FT}  
& \cellcolor{col33}67.4 
& \cellcolor{col33}71.0 
& \cellcolor{col33}74.0 \\

\midrule

\multirow{7}{*}{\rotatebox[origin=c]{90}{\makecell{SSFSL}}} 
&FixMatch \cite{sohn2020fixmatch} \textsubscript{NeurIPS'20}  
&\textcolor{Gray}{FT}
&39.3 
&49.9 
&57.2 \\

&\cellcolor{green!20}FixMatch w/ Temp. (ours)  
&\cellcolor{green!20}\textcolor{Gray}{FT}
&\cellcolor{green!20}57.7$^{\textcolor{Green}{+18.4}}$ 
&\cellcolor{green!20}65.7$^{\textcolor{Green}{+15.8}}$ 
&\cellcolor{green!20}71.2$^{\textcolor{Green}{+14.0}}$  \\
\cmidrule{2-6}
&DebiasPL \cite{wang2022debiased} \textsubscript{CVPR'22} 
&\textcolor{Gray}{FT}
&39.6
&49.8
&57.1 \\

&\cellcolor{green!20}DebiasPL w/ Temp. (ours) 
&\cellcolor{green!20}\textcolor{Gray}{FT} 
&\cellcolor{green!20}60.3$^{\textcolor{Green}{+20.7}}$ 
&\cellcolor{green!20}67.6$^{\textcolor{Green}{+17.8}}$ 
&\cellcolor{green!20}73.2$^{\textcolor{Green}{+16.1}}$  \\

\cmidrule{2-6}

& FineSSL \cite{gan2024erasing} \textsubscript{ICML'24} 
& \textcolor{Gray}{{PL}}  
& 57.6 
& 64.6  
& 68.9 \\

&\textbf{SWIFT (ours)}  
&\textcolor{Gray}{FT}
&\textbf{71.5$^{\textcolor{blue}{+4.1}}$}  
&\textbf{76.3$^{\textcolor{blue}{+5.3}}$}  
&\textbf{79.7$^{\textcolor{blue}{+5.7}}$}  \\

\midrule 
\multirow{2}{*}{\rotatebox[origin=c]{90}{Ref.}} 
&\cellcolor{gray!15}Fully Supervised  &\cellcolor{gray!15}\multirow{1}{*}{\textcolor{Gray}{FT}}  &\cellcolor{gray!15}74.8 &\cellcolor{gray!15}75.4 &\cellcolor{gray!15}76.0 \\

&\cellcolor{gray!15}Fully Supervised w/ RA  &\cellcolor{gray!15}\textcolor{Gray}{FT}  &\cellcolor{gray!15}76.0  &\cellcolor{gray!15}76.8  &\cellcolor{gray!15}77.2  \\

\bottomrule
\end{tabular}
    }
\end{minipage}
\hspace{\fill}
\begin{minipage}{0.38\textwidth}
\caption{ \small
{\bf SWIFT achieves SOTA SSFSL performance}.
We compare our methods with recent FSL and SSL methods that adapt the VLM OpenCLIP ViT-B/32 through prompt learning (PL), adapter learning (AL), and finetuning (FT).
Notably, our temperatures significantly improve FixMatch \cite{sohn2020fixmatch} and DebiasPL \cite{wang2022debiased} by 14-20\% accuracy.
Crucially, SWIFT outperforms 
SOTA FSL method SWAT \cite{liu2025few} that also exploits retrieval augmentation (RA) by 5 accuracy points (\textcolor{blue}{superscripts}). 
SWIFT even rivals \textcolor{gray}{fully supervised references}, which uses ground-truth labels for unlabeled data.  
     }
     \label{tab:compare_sota}
\end{minipage}
\vspace{-7mm}
\end{table}}

{\bf Compared Methods and Models.}
In addition to FixMatch and DebiasPL presented in Section~\ref{sec:problem_formulation_and_methods},
we compare our SWIFT with the state-of-the-art (SOTA) SSL method FineSSL \cite{gan2024erasing}, which adopts prompt learning with a frozen VLM.
We also compare with various FSL methods,
including popular ones that learn either a lightweight adapter \cite{clap24, lin2023multi, taskres, tipadapter, clipadapter, radford2021learning}
or prompts \cite{li2019learning, chen2022plot, wang2025attention} with a frozen VLM.
In particular, we compare SWIFT with the SOTA FSL methods that finetune a VLM's visual encoder, including FS-FT \cite{liu2025few} that finetunes on few-shot labeled data only, and SWAT \cite{liu2025few} that also retrieves open data to augment the training set.
For fair comparison with recent VLM-based FSL and SSL methods, we mainly study the open-source VLM OpenCLIP ViT-B/32 \cite{cherti2023reproducible}. We also demonstrate that SWIFT generalizes to other pretrained foundation models such as the DINOv2 ViT-B/14 \cite{oquab2024dinov}. 
Supplement \cref{sec:further_analysis} provides analysis of additional backbones, such as ResNet-50 and ViT, pretrained on ImageNet.
As a reference, we report results from supervised learning that finetunes the visual encoder on both labeled and unlabeled data (with ground-truth labels), optionally with retrieved data.

{\bf Implementation Details.}
For each dataset, we randomly sample 4-, 8-, and 16-shot labeled data from the official training set and repurpose the remaining training data as unlabeled data.
An exception is semi-Aves \cite{semi-aves}, where we use its official unlabeled in-domain data as the unlabeled set (details in Supplementary \cref{tab:datasets}). 
Following \cite{liu2025few, parashar2024neglected}, we retrieve 500 images per class from LAION-400M \cite{laion400m} for RA.
Adhering to the realistic ``validation-free protocol'' (\cref{sec:problem_formulation_and_methods}), we directly adopt hyperparameters (e.g., learning rate, weight decay, batch size) from \cite{liu2025few} for all datasets.
Notably, for temperatures, we follow \cite{su2021realistic} to search on semi-Aves, leading to the configurations: initializing $T_\text{loss}=0.07$ and learning it during VLM finetuning, and setting $T_\text{conf}=0.01$.
We apply them throughout experiments on all datasets.
All experiments are run on a single NVIDIA RTX 4090 (24GB) GPU with 50 GB disk space for data storage.

\begin{table}[t]
\centering
\small
\caption{\small 
\textbf{Ablation study of SWIFT.}
Each stage of SWIFT (\cref{fig:swift}) contributes significant accuracy gains when compared with directly applying  
FixMatch \cite{sohn2020fixmatch} to finetune the VLM OpenCLIP \cite{cherti2023reproducible}.
Notably, SWIFT also generalizes to the DINOv2 model, which is pretrained on large-scale unlabeled images using a self-supervised loss \cite{oquab2024dinov}.
\textcolor{Green}{Superscripts} mark the incremental improvements relative to the previous row.
We refer the reader to Supplementary \cref{tab:swift_dinov2_detail} and \cref{tab:ablate_swift_detail} for per-dataset analyses on the stage-wise design of SWIFT. 
}
\vspace{-3mm}
\label{tab:improvements_over_direct}
\setlength{\tabcolsep}{0.6mm}
\scalebox{0.85}{
\begin{tabular}{llllllll}
\toprule
 & & \multicolumn{6}{c}{mean accuracy over five datasets} \\
 \cmidrule(lr){3-8}
\multirow{2}{*}{method} & \multirow{2}{*}{\makecell{training\\data}} & \multicolumn{3}{c}{OpenCLIP ViT-B/32}  &  \multicolumn{3}{c}{DINOv2 ViT-B/14} \\
\cmidrule(lr){3-5}  \cmidrule(lr){6-8}
&  &4-shot &8-shot &16-shot  &4-shot &8-shot &16-shot \\
\midrule

\rowcolor{gray!15}
FS-FT \cite{liu2025few} \textsubscript{CVPR'25} & L & 61.0 & 65.8 & 69.1 &56.5  &71.1  &79.6 \\
\midrule

FixMatch\cite{sohn2020fixmatch} direct & L+U 
& 39.3 
& 49.9 
& 57.2 
&50.2
&66.3  
&77.1 \\

\enspace + stage-1: Classifier Init.  & L+U
& 56.3$^{\textcolor{Green}{+17.0}}$ 
& 59.8$^{\textcolor{Green}{+9.9}}$ 
& 62.2$^{\textcolor{Green}{+5.0}}$ 
& 56.7$^{\textcolor{Green}{+6.5}}$ 
& 71.5$^{\textcolor{Green}{+5.2}}$ 
& 80.3$^{\textcolor{Green}{+3.1}}$\\

\enspace \enspace+ stage-2: Semi-Sup. FT & L+U+R 
& 68.8$^{\textcolor{Green}{+11.1}}$ 
& 73.3$^{\textcolor{Green}{+7.6}}$ 
& 77.3$^{\textcolor{Green}{+6.1}}$ 
& 76.3$^{\textcolor{Green}{+19.6}}$ 
& 82.0$^{\textcolor{Green}{+10.5}}$ 
& 86.3$^{\textcolor{Green}{+6.0}}$\\

\rowcolor{green!20} \enspace \enspace \enspace + stage-3: FS-FT (SWIFT) & L+U+R 
& \textbf{71.5$^{\textcolor{Green}{+2.7}}$} 
& \textbf{76.3$^{\textcolor{Green}{+3.0}}$} 
& \textbf{79.7$^{\textcolor{Green}{+2.4}}$} 
& \textbf{78.2$^{\textcolor{Green}{+1.9}}$} 
& \textbf{84.4$^{\textcolor{Green}{+2.4}}$} 
& \textbf{87.8$^{\textcolor{Green}{+1.5}}$}
\\

\bottomrule

\end{tabular}}
\vspace{-4mm}
\end{table}

\subsection{Experimental Results}
{\em SWIFT achieves state-of-the-art SSFSL performance.}
As shown in \cref{tab:compare_sota}, our simple temperature-based remedy significantly improves existing SSL methods, e.g., FixMatch \cite{sohn2020fixmatch} and DebiasPL \cite{wang2022debiased}, enabling successful VLM finetuning.
Our improved versions even surpass the SOTA SSL method FineSSL \cite{gan2024erasing}, reinforcing our motivation that finetuning VLM performs better than prompting a frozen VLM.
Importantly, SWIFT outperforms the SOTA FSL method SWAT \cite{liu2025few}, which also exploits retrieved open data for VLM finetuning. The strong accuracy gains demonstrate our successful exploitation of unlabeled data by using temperatures.
Furthermore, SWIFT even rivals fully supervised learning, which finetunes VLM using few-shot labeled data, unlabeled data (with ground-truth labels), and retrieved open data in a single training stage. The results highlight the advantages of SWIFT's multi-stage training pipeline.

{\em Ablation study validates the generalization of SWIFT.}
\cref{tab:improvements_over_direct} highlights the significant progressive performance gains by each stage in SWIFT.
Importantly, SWIFT not only improves VLM finetuning, but also generalizes to self-supervised pretrained vision foundation models, e.g., DINOv2 \cite{oquab2024dinov}, achieving even better SSFSL performance.
In addition, we show in the Supplement \cref{tab:swift_debiaspl} that SWIFT also improves the stronger SSL method, DebiasPL \cite{wang2022debiased}, serving as a plug-and-play framework for improving existing SSL methods.

\begin{figure}[t]
\centering
\includegraphics[width=\textwidth, clip=true,trim = 0mm 0mm 0mm 0mm]{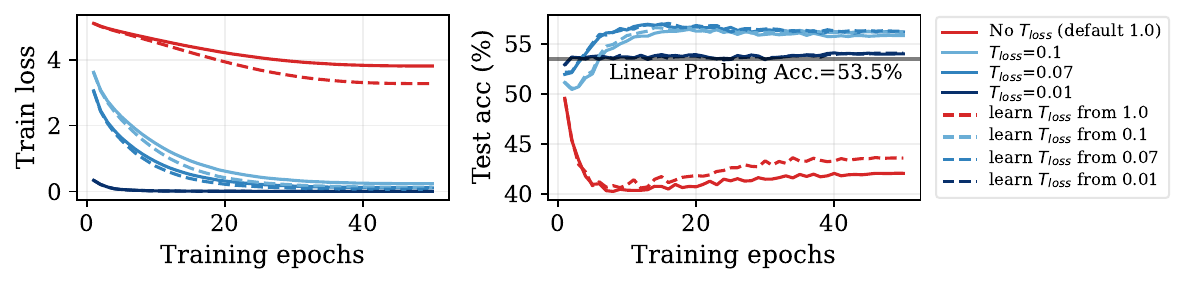}
\vspace{-6mm}
\caption{\small
{\bf Loss temperature $T_{\text{loss}}$ strengthens training supervision.}
We compare the training loss and test accuracy under different settings of $T_{\text{loss}}$ when finetuning the OpenCLIP on 16-shot labeled examples from the semi-Aves dataset \cite{semi-aves}.
Notably, compared to finetuning without temperature (default value of 1.0), using a small loss temperature (e.g., 0.1 or 0.07) effectively accelerates the training loss reduction, improving test accuracy over the few-shot linear probing baseline.
In addition, learning the $T_{\text{loss}}$ dynamically during the training (dashed lines) outperforms using a fixed $T_{\text{loss}}$ (solid lines), due to increased flexibility.
Supplementary \cref{fig:acc_w_varying_Tloss} studies learning $T_{\text{loss}}$ from various initial values with three random seeds, confirming our observations.
}
\label{fig:loss_temp_fixmatch} 
\vspace{-3mm}
\end{figure}

{\em Using a small loss temperature $T_{\text{loss}}$ enables effective VLM finetuning.}
\cref{fig:loss_temp_fixmatch} studies the effect of loss temperature when finetuning the VLM on few-shot labeled data.
The results clearly show that adopting $T_{\text{loss}}$ significantly accelerates the reduction of the training loss and improves test accuracy.
Supplementary \cref{tab:ablate_swift_detail} provides per-dataset
accuracy improvements,
confirming the robustness of our proposed temperature-based solution.

\begin{figure}[t]
\centering
\includegraphics[width=\textwidth, clip=true,trim = 15mm 45mm 15mm 45mm]{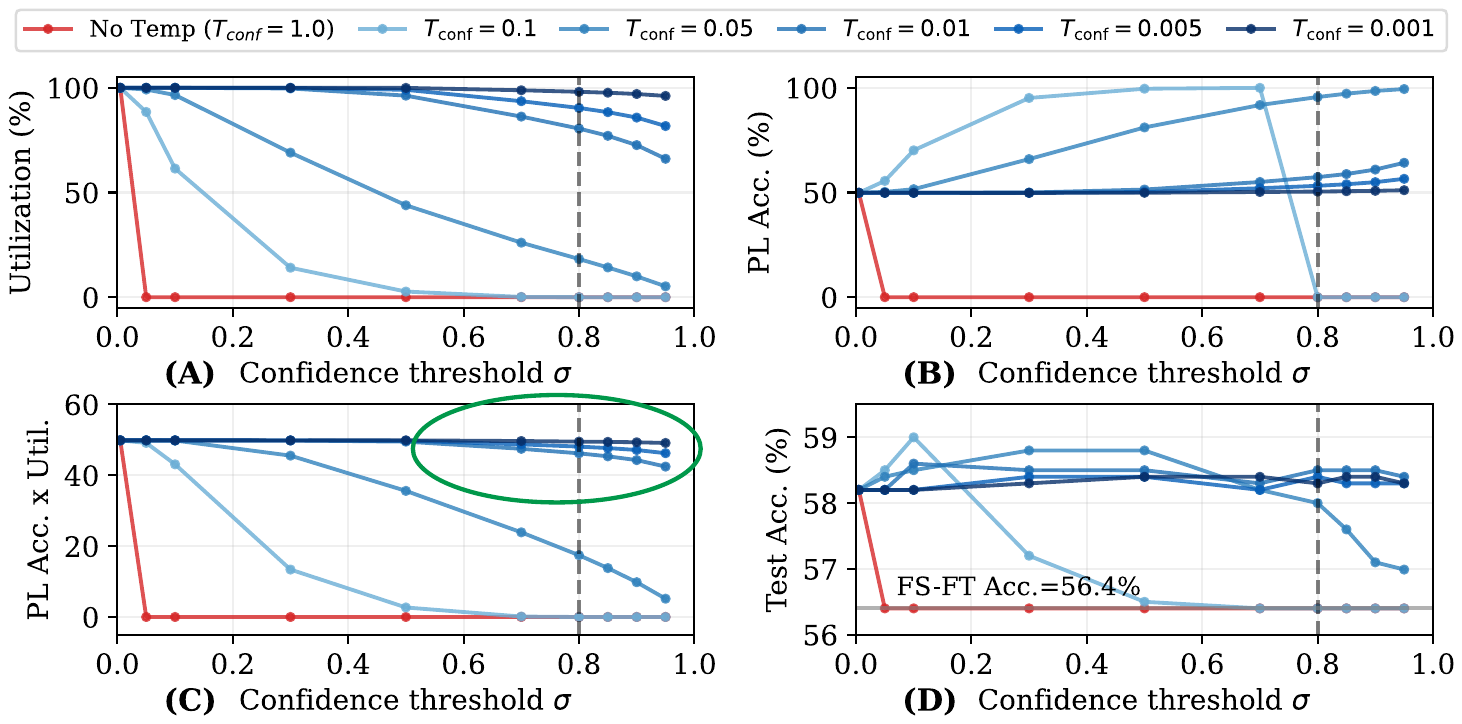}
\vspace{-6mm}
\caption{\small
{\bf Confidence temperature $T_{\text{conf}}$ enables robust utilization of unlabeled data.}
Building on the $T_{\text{loss}}$ in \cref{fig:loss_temp_fixmatch},
we experiment with FixMatch on OpenCLIP with 16-shot data from the semi-Aves dataset under various $T_{\text{conf}}$ settings.
\textbf{(A)}: small $T_{\text{conf}}$ consistently improves utilization rate (percentage of selected pseudo-labels with confidence $>$ a threshold $\sigma$, averaged over 50 training epochs); 
\textbf{(B)}: the pseudo-label accuracy increases with a large confidence threshold and small temperatures; 
\textbf{(C)} shows the product of the utilization rate and pseudo-label accuracy, 
highlighting the benefits of using small $T_{\text{conf}}$ for selecting a greater quantity of correct pseudo-labels for SSL (cf. \textcolor{Green}{green circle});
\textbf{(D)}: setting the $T_{\text{conf}}$ in a broad range within $[0.001, 0.05]$ significantly improves test accuracy over the FS-FT baseline \cite{liu2025few}.
Importantly, using a small $T_{\text{conf}}$ (e.g., 0.01) yields consistently strong accuracy gains across various thresholds.
}
\label{fig:confidence_temp_fixmatch}
\end{figure}

{\em Using a confidence temperature $T_{\text{conf}}$ enables robust utilization of unlabeled data.}
\cref{fig:confidence_temp_fixmatch} studies the utilization rate, pseudo-label accuracy, and test accuracy when finetuning VLM using FixMatch under various confidence temperatures.
The results highlight the cruciality of tuning the confidence threshold $\sigma$ for utilizing unlabeled data when no confidence temperature is used (red curves), which is impractical in the realistic ``validation-free''  (\cref{fig:problem-setup}).
In contrast, confidence temperature provides a robust solution, as evidenced by the consistent and strong accuracy gains when setting $T_{\text{conf}}$ across a broad range (darker blue lines).
The Supplementary \cref{tab:ablate_swift_detail} provides more detailed results on each dataset, further validating the robustness of using temperatures.

{\em 
Temperature enables learning a better classifier, which improves subsequent semi-supervised finetuning.}
\cref{tab:cls_stage1} compares different classifier initialization for stage-1 and stage-2 training of SWIFT. 
Results show that, while using text embedding \cite{parashar2024neglected} outperforms random initialization \cite{zhang2025revisiting, sohn2020fixmatch}, further adding our loss temperature notably improves performance.
In addition, our SWIFT adopts the stage-1 learned classifier for the subsequent stage-2 finetuning, achieving better accuracy than using classifier initialized with text-embedding.
Our results are similar to the observations in \cite{kumar2022finetuning}, which find that learning the classifier improves robust VLM finetuning, despite they did not explore temperatures.

{\em Further analysis.}
We provide additional in-depth analyses 
in the Supplement. Below, we summarize our key findings.
\cref{fig:vit_confidence_comparison} and \cref{fig:rn50_confidence_comparison} compare the distribution of confidence scores of more pretrained models (ImageNet-pretrained vs. OpenCLIP and CLIP), and different architectures (ResNet vs. ViT). The results confirm that the observed ``flat softmax probabilities'' arise from the contrastive pretraining objective rather than from architectural differences.
\cref{fig:softmax_distribution_stage1} visualizes the distribution of softmax probabilities after stage-1 training, confirming that logits remain flat and temperatures are needed for subsequent semi-supervised finetuning. 
In addition, \cref{fig:small_threshold} studies using a small confidence threshold rather than using a confidence temperature, and \cref{fig:freematch} further experiments the adaptive threshold SSL method, FreeMatch \cite{wang2023freematch}. The results show that while these methods improve the utilization of unlabeled data, they yield poor test accuracy due to the lack of training supervision, highlighting the importance of using a loss temperature.
\cref{fig:acc_w_varying_Tloss} compares different initial value for learning the loss temperature $T_{\text{loss}}$. The results show that using a small $T_{\text{loss}}$ yields consistent and strong accuracy gains with small standard deviations across three random seeds. 
\cref{tab:ablate_ts_probing} and \cref{tab:learn_temp_fsft_fixmatch} extensively study the effect of loss temperature across different learning paradigms (linear probing, few-shot and semi-supervised finetuning), different pretrained models (ImageNet-pretrained and VLMs), and different model architectures (ResNet vs. ViT), validating the importance of loss temperature when adapting a VLM across all settings.
Moreover, \cref{tab:ablate_retrieved_images} studies the impact of the number of retrieved images, demonstrating improved performance with more retrieved data.

\section{Impacts, Limitations, and Future Work}
\label{sec:discussions}

{\bf Broad Impacts.}
SSFSL has promising real-world applications, and our work provides a simple yet effective solution by leveraging open-source pretrained foundation models and open data. 
However, we acknowledge potential societal risks. First, using retrieved open data may cause the finetuned model to overgeneralize to open-set or anomalous inputs. Second, as with other work that builds on foundation models, our approach may inherit biases from the pretrained VLM, potentially raising fairness concerns.

{\bf Limitations and Future Work.}
While our work advances SSFSL within a realistic setup and resolves the failures of established SSL methods in VLM finetuning, several avenues for improvement remain. 
First, 
although the validation-free setup realistically simulates real-world scenarios where scarce labeled data do not support validation,
future research could explore unlabeled and retrieved data to aid validation so to allow hyperparameter tuning for specific tasks.
Moreover, while the datasets used are standard in the literature, 
they may not fully reflect the extreme natural class imbalances encountered in real-world applications. Future work could construct more diverse datasets to facilitate SSFSL research.

{
\setlength{\tabcolsep}{0.2em} 
\begin{table*}[t]
  \centering
    \caption{\small
    \textbf{Comparison of classifier initialization methods for stage-1 linear probing and stage-2 semi-supervised finetuning (SS-FT) in SWIFT.} 
    We experiment with the VLM OpenCLIP ViT-B/32 across five datasets.
    \textbf{Left:} compared to prior linear probing methods that either use random initialization \cite{radford2021learning} or text embedding \cite{parashar2024neglected} to initialize the classifier weights, our approach, which adopts a learnable $T_\text{loss}$ initialized to 0.07, achieves 2-4\% average accuracy gains. The results highlight the benefits of using temperature when adapting VLMs. 
    \textbf{Right:} 
    In contrast to prior SSFSL methods that either randomly initialize the classifier weights \cite{zhang2025revisiting, sohn2020fixmatch} or use text embeddings \cite{liu2025few}, SWIFT uses the stage-1 learned classifier for semi-supervised finetuning, achieving over 2\% accuracy gains, especially in lower shots.
    The results validate our stage-wise design.
    \textcolor{Green}{Superscripts} denote improvements over the text embedding.
    }
    \vspace{-2mm}
  \begin{subtable}[t]{0.49\textwidth}
  \centering
  \scalebox{0.75}{
\begin{tabular}{cllll}
\toprule

& \multirow{2}{*}{\makecell{stage-1\\classifier init.}} 
& \multicolumn{3}{c}{mean acc. over five datasets} \\
\cmidrule(lr){3-5}
& &4-shot &8-shot &16-shot \\
\midrule

\multirow{3}{*}{\rotatebox[origin=c]{90}{\makecell{Linear\\probing}}} 
&random \cite{radford2021learning} &22.5 &36.4 &47.3 \\

& text \cite{parashar2024neglected} &54.9 &57.6 &59.9 \\

&\cellcolor{green!20}{text + $T_{\text{loss}}$ (ours)}  
&\cellcolor{green!20}\textbf{57.0$^{\textcolor{Green}{+2.1}}$} 
&\cellcolor{green!20}\textbf{61.2$^{\textcolor{Green}{+3.6}}$} 
&\cellcolor{green!20}\textbf{64.7$^{\textcolor{Green}{+4.8}}$} \\
\bottomrule

\end{tabular}}
\label{tab:cls_stage1}
\end{subtable}
\hfill
  \begin{subtable}[t]{0.49\textwidth}
  \centering
  \scalebox{0.75}{
\begin{tabular}{cllll}
\toprule

&\multirow{2}{*}{\makecell{stage-2\\classifier init.}} & \multicolumn{3}{c}{mean acc. over five datasets} \\
\cmidrule{3-5}
& &4-shot &8-shot &16-shot \\
\midrule

\multirow{3}{*}{\rotatebox[origin=c]{90}{\makecell{SS-FT}}} 

& random \cite{zhang2025revisiting}
& 65.3
& 60.9
& 66.8 \\

&text \cite{liu2025few} 
& 66.0
& 71.4 
& 77.0 \\

& \cellcolor{green!20}stage-1 learned (ours)
& \cellcolor{green!20}\textbf{68.8$^{\textcolor{Green}{+2.8}}$} 
& \cellcolor{green!20}\textbf{73.3$^{\textcolor{Green}{+1.9}}$} 
& \cellcolor{green!20}\textbf{77.3$^{\textcolor{Green}{+0.3}}$}\\

\bottomrule
    \end{tabular}}
    \label{tab:cls_stage2}
  \end{subtable}
  \label{tab:classifier_init}
  \vspace{-2mm}
\end{table*}
}

\section{Conclusions}
\label{sec:conclusions}

We study semi-supervised few-shot learning (SSFSL) from a realistic ``auto-annotation'' perspective, which motivates us to exploit VLMs and open data to improve performance.
Our extensive experiments reveal that representative SSL methods fail to finetune VLM effectively.
We identify the root cause of failures: VLMs produce a ``flat'' distribution of softmax probabilities, resulting in zero utilization of unlabeled data and weak training supervision that hinders effective finetuning.
To address these challenges, we propose a simple technique that uses temperatures to sharpen softmax output.
In addition, we further exploit retrieved open data to boost SSFSL performance.
Building on these insights, we present a simple SSFSL method, SWIFT, which effectively finetunes a VLM with stage-wise training. 
Extensive experiments demonstrate that SWIFT significantly outperforms existing methods by 5 average accuracy points, establishing a new state-of-the-art across five benchmarks.

\section*{Acknowledgements}
This work was supported by the Science and Technology Development Fund of Macau (0058/2025/RIA2, 0067/2024/ITP2), the University of Macau (SRG2023-00044-FST), the Institute of Collaborative Innovation, and CK Foundation.
The authors thank Pardis Taghavi for discussions, the CSE Department at Texas A\&M University, and the advanced computing resources and consultation provided by Texas A\&M High Performance Research Computing (HPRC).
Part of this work used the Delta system at the National Center for Supercomputing Applications [award OAC 2005572] through allocation [CIS250837, CIS250928] from the Advanced Cyberinfrastructure Coordination Ecosystem: Services \& Support (ACCESS) program, which is supported by National Science Foundation grants \#2138259, \#2138286, \#2138307, \#2137603, and \#2138296.

%
%
\bibliographystyle{splncs04}
\bibliography{main}

\clearpage

{
   \newpage

        \centering
        \Large
        \textbf{Solving Semi-Supervised Few-Shot Learning from an Auto-Annotation Perspective} \\ \vspace{0.5em} {(Supplementary Material)}\\
        \large
   }

\renewcommand{\thesection}{\Alph{section}}
\renewcommand{\theHsection}{\Alph{section}}
\setcounter{section}{0} 

\section*{}
\begin{center}
    \emph{\bf \em \large Outline}
\end{center}

This document supplements our main paper with detailed results and comprehensive analyses. It is organized as below: 

\begin{itemize}
\item {\bf Section \ref{sec:datasets}} 
describes benchmarking datasets and shows examples of few-shot labeled, unlabeled, and retrieved noisy labeled data.

\item {\bf Section \ref{sec:hyperparameters}} provides details of the hyperparameters in our experiments.

\item {\bf Section \ref{sec:further_analysis}} provides more visualizations of softmax probabilities across different pretrained backbones, and more analysis on temperatures. 

\item {\bf Section \ref{sec:details}} reports per-dataset benchmarking and ablation results.

\item {\bf Section \ref{sec:pseudocode}} provides pseudo-code of our teperature-based solution with FixMatch.

\item {\bf Section \ref{sec:Demo-code}}  provides code and instructions for reproducing our results.

\end{itemize}

\section{Summary of Datasets}
\label{sec:datasets}
\begin{figure*}[t]
  \centering
  \small
  \includegraphics[width=.99\linewidth, clip=true, trim = 0mm 0mm 0mm 0mm]{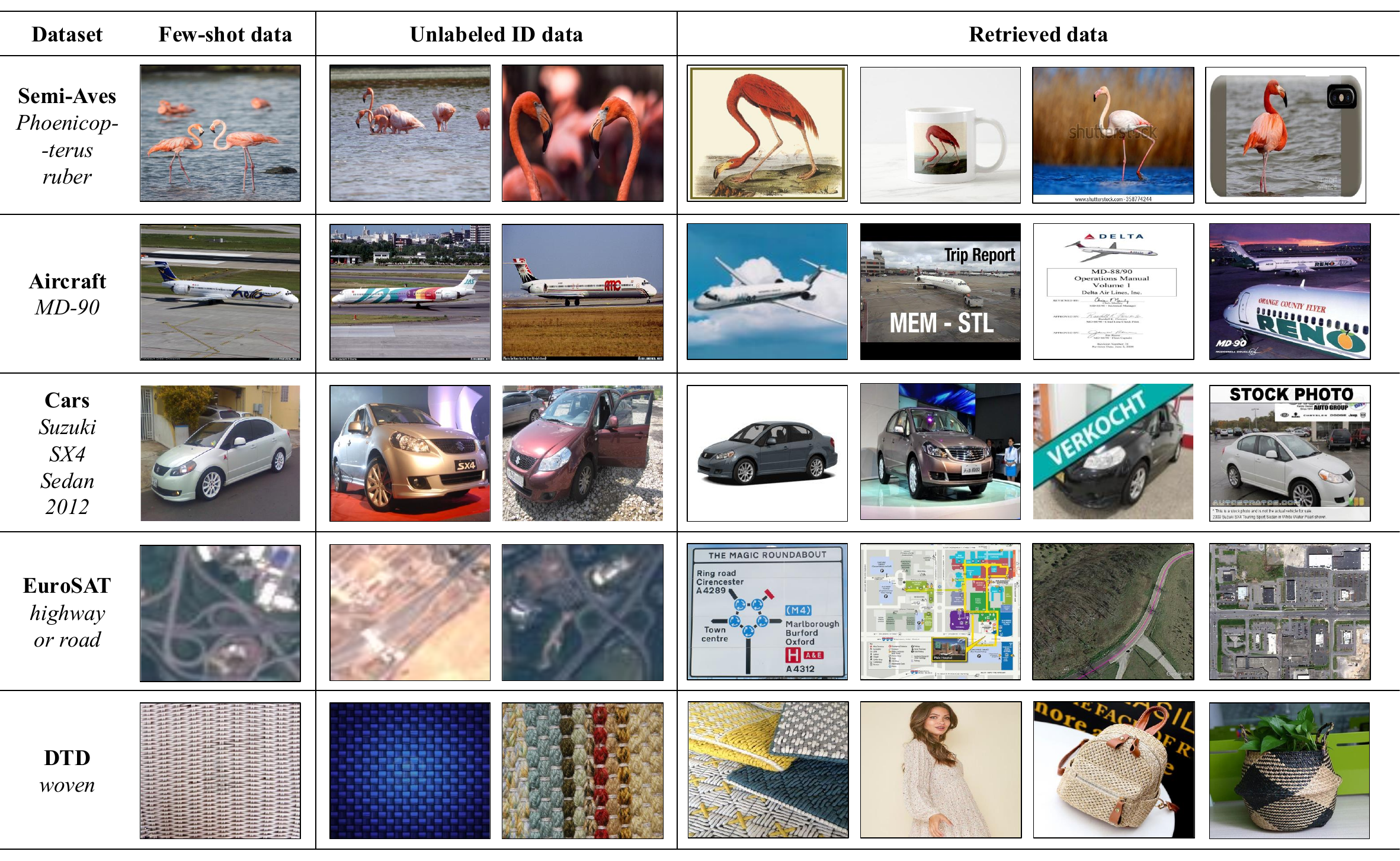}
  \vspace{-3mm}
  \caption{\small
  \textbf{Visual examples} of few-shot labeled data, unlabeled data, and retrieved data for representative classes from each dataset.
  Notably, the unlabeled data follows the same distribution as the few-shot labeled data, exhibiting similar visual appearances. 
  In contrast, the retrieved data demonstrates domain gaps from the task-specific data \cite{liu2025few}, presenting different visual patterns in styles, backgrounds, and resolutions. 
  }
  \vspace{-1mm}
  \label{fig:example_data}
\end{figure*}

\cref{tab:datasets} presents a detailed summary of the five fine-grained datasets used in our experiments. 
We sample few-shot labeled data from the official training set following \cite{liu2025few, clap24}, and repurpose the remaining training and validation images as the unlabeled data.
The only exception is semi-Aves \cite{semi-aves}, where we use its official unlabeled in-domain data as the unlabeled data.
We follow \cite{parashar2024neglected, liu2025few} to retrieve data from OpenCLIP's publicly available dataset LAION-400M (under CC-BY 4.0 License) \cite{laion400m}. Specifically, 
we conduct ``string-matching'' to retrieve pretraining images whose captions contain any synonyms of downstream concepts. 
\cite{parashar2024neglected} shows that string-matching-based retrieval improves both efficiency and diversity.
We then sample 500 images for each class following \cite{liu2025few}. 
For classes with fewer than 500 images, we use all the retrieved data.
We include the number of retrieved data in \cref{tab:datasets} and 
present visual examples in \cref{fig:example_data}.

{
\setlength{\tabcolsep}{1.5em} 
\begin{table}[ht]
\centering
\small
\caption{ \textbf{Details of five fine-grained benchmarks.}
We list the number of images in the official training, validation, test, and unlabeled ID sets for each dataset.
We sample $K$-shot ($K=4, 8, 16$) labeled data from the official training set, and then repurpose the remaining training and validation images as the unlabeled data. 
}
\vspace{-3mm}
\scalebox{0.9}{
\begin{tabular}{llll}
\toprule 
dataset & \# cls &official train/val/test/unlabeled & retrieved \\
\midrule
semi-Aves~\cite{semi-aves} & 200 & 3,959 / 2,000 / 4,000 / 26,640   &47,006  \\
Aircraft~\cite{aircraft} & 100 & 3,334 / 3,333 / 3,333 / NA &30,429  \\
Stanford Cars~\cite{cars} & 196 & 6,509 / 1,635 / 8,041 / NA &80,648 \\
EuroSAT~\cite{eurosat} & 10 & 13,500 / 5,400 / 8,100 / NA &1,871 \\
DTD~\cite{dtd} & 47 & 2,820 / 1,128 / 1,692 / NA &23,364  \\
\bottomrule
\end{tabular}}
\label{tab:datasets}
\vspace{-3mm}
\end{table}
}

{\setlength{\tabcolsep}{0.9em}
\begin{table}[t]
\centering
\caption{ \textbf{Comparison of computational cost between SWIFT and baselines.}
We experiment with all methods on the OpenCLIP model using the 16-shot setting on the semi-Aves dataset by training for 50 epochs on an NVIDIA RTX 4090 GPU.
As SWIFT uses more data, it takes more training time. Importantly, it significantly outperforms the compared methods in terms of test accuracy.
}
\label{tab:compare_compute_cost}
\vspace{-3mm}    
\scalebox{0.9}{
\begin{tabular}{llrcc}
\toprule
 & training data & data storage & training time & test acc \\ 
 \midrule
FS-FT \cite{liu2025few} & L & 0.6 GB & 0.3 hr & 56.5 \\ 
FixMatch w/ VLM \cite{sohn2020fixmatch} & L+U & 5.8 GB & 1.2 hr & 31.8 \\ 
SWAT \cite{liu2025few} & L+R & 9.7 GB & 2.0 hr  & 63.1 \\
\textbf{SWIFT (ours)} & L+R+U & 14.9 GB & 8.0 hr & \textbf{68.7} \\ 
\bottomrule
\end{tabular}}
\vspace{-3mm}
\end{table}}

\section{Hyperparameters}
\label{sec:hyperparameters}

SWIFT finetunes a VLM's visual encoder in three stages, leveraging few-shot labeled and abundant unlabeled data, along with a large amount of retrieved data with noisy labels 
Specifically, following our realistic SSFSL setup that eschews a validation set, we follow \cite{liu2025few} to directly adopt the hyperparameters reported in the literature to our SWIFT experiments across all datasets.
Below, we list the hyperparameters in each stage.

\textbf{Stage-1 Classifier Initialization.}
We initialize the classifier weights with text embeddings of the $C$ class names \cite{parashar2024neglected} and further learn the classifier on few-shot labeled data. 
We train for 50 epochs with a learning rate of 1e-4, a weight decay of 1e-2, an \emph{AdamW} optimizer, a batch size of 32, and a cosine-annealed learning rate scheduler.

\textbf{Stage-2 Semi-Supervised Finetuning.}
SWIFT can be integrated with various existing SSL methods, such as FixMatch \cite{sohn2020fixmatch} and DebiasPL \cite{wang2022debiased}.
Specifically, we follow \cite{liu2025few} to mix the retrieved data (with noisy labels) and few-shot labeled data in a batch for calculating the cross-entropy loss $\mathcal{L}_l$.
Additionally, following established practice \cite{sohn2020fixmatch}, we add few-shot data (with labels removed) to the unlabeled data to augment the unlabeled set for calculating the cross entropy loss $\mathcal{L}_u$.
We do not add retrieved data to the unlabeled set, as it has domain gaps relative to the task-specific data \cite{liu2025few}.
We follow \cite{liu2025few} to use a learning rate of 1e-4 to update the classifier, and a smaller learning rate of 1e-6 to update the visual encoder to preserve the pretrained features, with a cosine annealing learning rate scheduler. 
The weight decay is set to 1e-2 with the \emph{AdamW} optimizer.
For the FixMatch loss calculation, we use a batch size of 32 for labeled data and a multiplier $\mu$ of 5 for unlabeled data, yielding a total of 192 training samples per training batch.
We use a fixed confidence threshold of 0.8 with a confidence temperature $T_\text{conf} = 0.01$ and set the loss temperature to $T_\text{loss}$ to a learnable parameter initialized to 0.07.
Same as stage-1, we train for 50 epochs.

\textbf{Stage-3 Few-shot Finetuning.}
To mitigate domain gaps and imbalanced distributions in the retrieved data \cite{liu2025few}, we finetune the model on few-shot labeled data for 10 epochs, using the same hyperparameters as in stage-2 training.
For all three stages, we learn a loss temperature $T_\text{loss}$ initialized to 0.07 with a learning rate of 1e-4 and a weight decay of 1e-2.

\textbf{Other FSL and SSL baselines.}
We obtain FSL baseline results directly from \cite{liu2025few}. For SSL methods, including FixMatch \cite{sohn2020fixmatch}, DebiasPL \cite{wang2022debiased}, and FineSSL \cite{gan2024erasing}, we directly adopt their reported hyperparameters and run them on our datasets.
\cref{tab:compare_compute_cost} compares the compute cost between SWIFT and the baselines.

\section{Further Analyses}
\label{sec:further_analysis}
In this section, we provide further analyses by comparing more temperature settings across different learning strategies and pretrained backbones.

\begin{figure}[t]
\centering
\begin{minipage}[t]{0.48\columnwidth}
\includegraphics[width=1.0 \textwidth, clip=true,trim = 0mm 3mm 0mm 0mm]{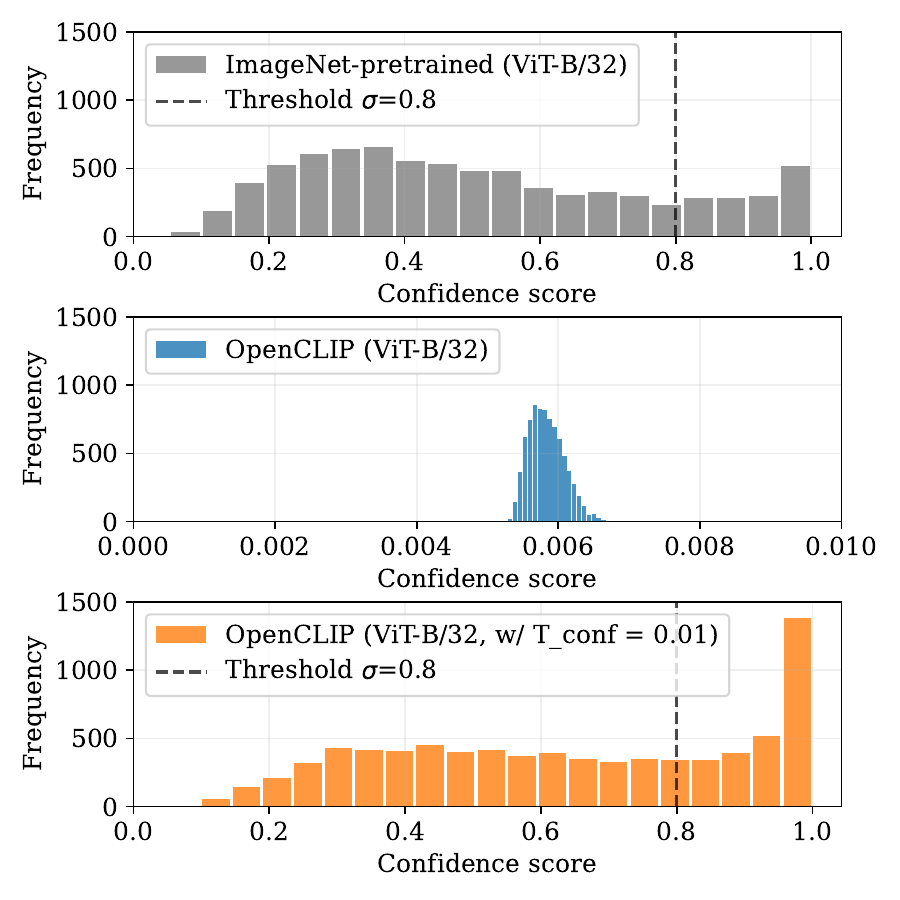}
\vspace{-5mm}
\caption{\small 
\textbf{Comparison of the confidence scores between ImageNet-pretrained and OpenCLIP ViT-B/32 models.}
For the ImageNet-pretrained backbone, we run linear probing with a randomly initialized classifier on 16-shot labeled data from the semi-Aves dataset \cite{semi-aves}.
For the OpenCLIP backbone \cite{cherti2023reproducible}, we use the text embeddings of class names for initializing the classifier weights.
}
\label{fig:vit_confidence_comparison}
    \end{minipage}\hfill
    \begin{minipage}[t]{0.48\columnwidth}
\centering
\includegraphics[width=1.0\textwidth, clip=true,trim = 0mm 3mm 0mm 0mm]{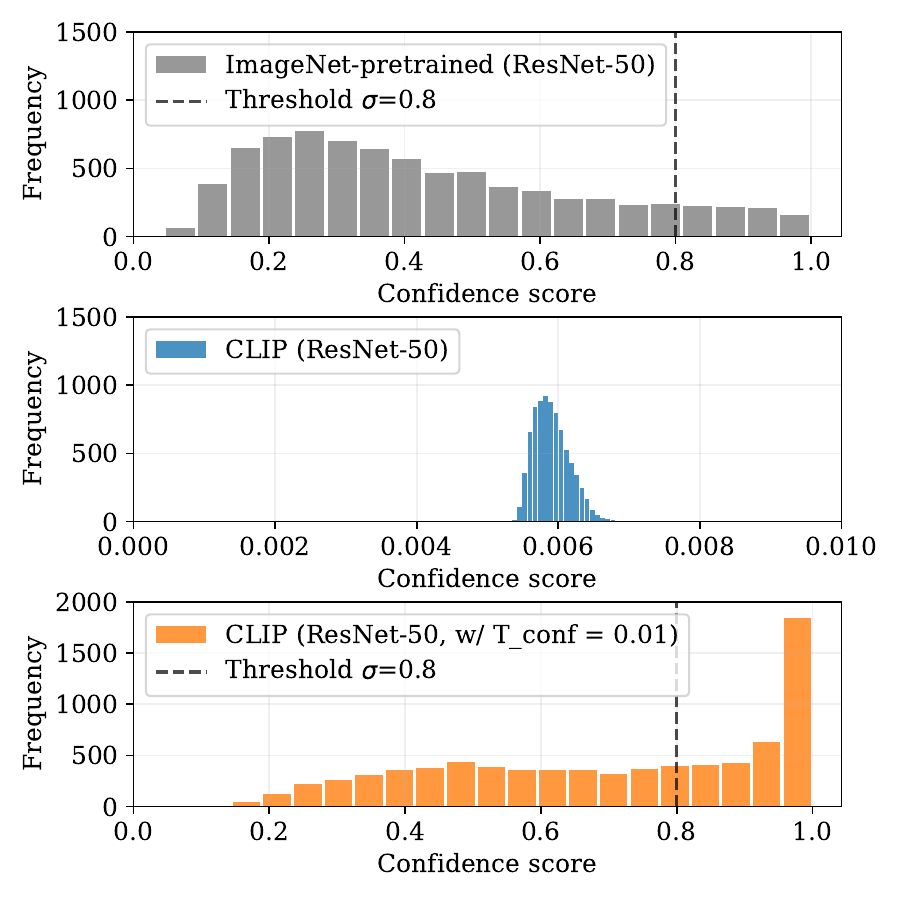}
\vspace{-5mm}
\caption{\small 
\textbf{Comparison of the confidence between ImageNet-pretrained and CLIP ResNet-50 models.}
Following the same settings in \cref{fig:vit_confidence_comparison}, we compare the confidence scores for the ResNet-50 backbone \cite{he2016deep}.
The small confidence scores from the CLIP ResNet-50 model \cite{radford2021learning} confirms that
contrastive pretraining loss is the cause of small softmax confidences, rather than architectural differences.
}
\label{fig:rn50_confidence_comparison}
    \end{minipage}
\vspace{-7mm}
\end{figure}

\textbf{Comparison of confidence scores across more architectures.}
\cref{fig:vit_confidence_comparison} and \cref{fig:rn50_confidence_comparison} compare the confidence scores between ImageNet-pretrained backbones and the VLM CLIP backbones. 
Results clearly show that CLIP models, regardless of ViT or ResNet architecture, produce small confidence scores that result in zero utilization of unlabeled data for SSFSL.
This confirms that the ``flat softmax probabilities'' result from contrastive pretraining loss rather than architecture differences.
Using a confidence temperature $T_\text{conf}=0.01$ increases both CLIP ViT and ResNet models' confidences, enabling the utilization of pseudo-labeled data with a default high confidence threshold (e.g., 0.8).

\begin{figure}[h]
\centering
\includegraphics[width=\textwidth, clip=true, trim = 0mm 0mm 0mm 0mm]{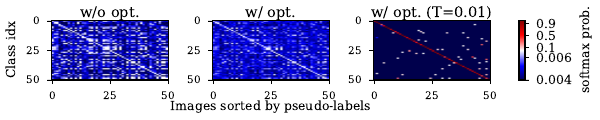}
\vspace{-7mm}
\caption{\small
{\bf Visualization of softmax distribution after stage-1 training.}
Using the same data setting as \cref{fig:diagnosis}, we show the softmax distribution for stage-1 linear probing using text embedding to initialize classifier weights (left, ``w/o opt''), using text embedding with learnable $T_\text{loss}$ (middle, ``w/ opt''), and further using a temperature of 0.01 to sharpen the softmax distribution (right).
We show the first 50 classes of semi-Aves for better visualization.
Results clearly show that stage-1 training does not solve the ``flat softmax distribution'' issue, where applying our temperature does.
}
\label{fig:softmax_distribution_stage1}
\vspace{-3mm}
\end{figure}

\textbf{Visualization of softmax distribution after stage-1 training.}
\cref{fig:softmax_distribution_stage1} shows the softmax distribution after stage 1, which remains flat. The results confirm the necessity of using temperatures in the subsequent SSL finetuning.

\begin{figure}[t]
\centering
\includegraphics[width=\textwidth, clip=true, trim = 0mm 0mm 0mm 0mm]{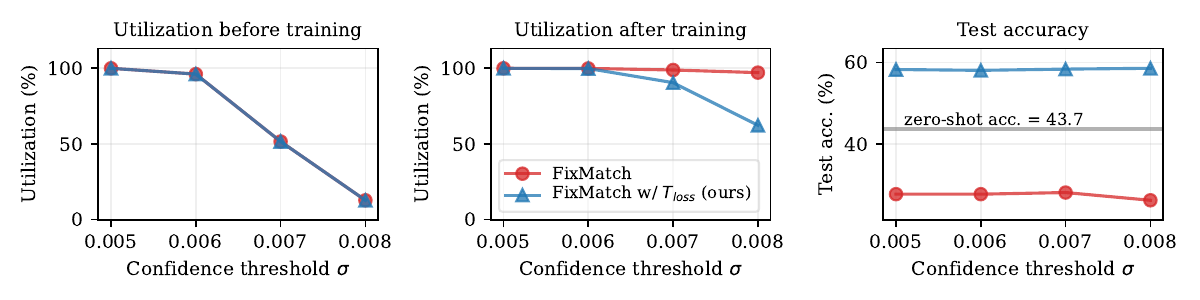}
\vspace{-7mm}
\caption{\small
{\bf Impact of small confidence thresholds.}
We experiment with FixMatch on OpenCLIP with 16-shot data on the semi-Aves dataset using small confidence thresholds. We report the utilization before and after training, and the final test accuracy under various confidence thresholds. With small thresholds, FixMatch does increase the utilization of unlabeled data but still yields poor accuracy.
This is due to the weak training supervision.
Instead, applying our $T_\text{loss}$ strengthens the training supervision and helps FixMatch boost performance (ref. the right plot and also \cref{fig:diagnosis} (C)). 
}
\label{fig:small_threshold}
\vspace{-3mm}
\end{figure}

\begin{figure}[t]
\centering
\includegraphics[width=\textwidth, clip=true, trim = 2mm 0mm 2mm 0mm]{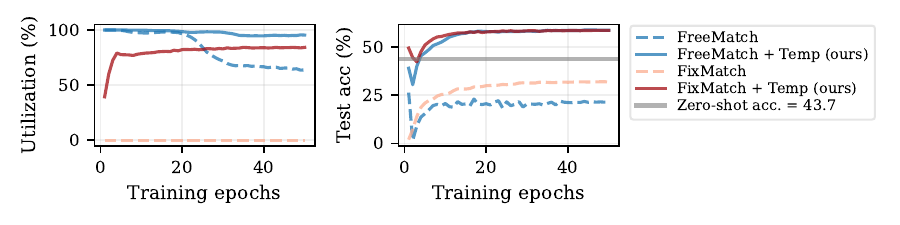}
\vspace{-9mm}
\caption{\small
{\bf Comparison of utilization rate and test accuracy between FixMatch and FreeMatch.}
Using the same setting of \cref{fig:small_threshold}, we study FreeMatch, which adaptively adjusts per-class threshold based on the model's learning status.
Results show that while FreeMatch yields $>$60\% utilization, it yields low accuracy, due to the lack of training supervision. However, adding $T_\text{loss}$ effectively increases test accuracy.
}
\label{fig:freematch}
\end{figure}

\textbf{Analysis of adaptive confidence threshold.}
Instead of using a small confidence temperature along with a fixed high confidence threshold for selecting high-quality pseudo-labeled data, one may wonder what if we do not use confidence temperature but choose a threshold value that reflects the VLM's softmax output values.
\cref{fig:small_threshold} shows that using a small confidence threshold does increase the utilization rate, but still leads to poor accuracy. This is due to the weak training supervision. However, applying our loss temperature $T_\text{loss}$ strengthens the training supervision and helps FixMatch boost performance (\cref{fig:diagnosis} (C)). 
The results highlight the importance of temperatures for effective finetuning of VLMs. We further study a recent adaptive thresholding method, FreeMatch \cite{wang2023freematch}, which yields similar results (\cref{fig:freematch}).

\begin{figure}[!t]
\centering
\begin{minipage}{0.53\columnwidth}
    \centering
\includegraphics[width=1.0 \textwidth, clip=true,trim = 0mm 0mm 0mm 0mm]{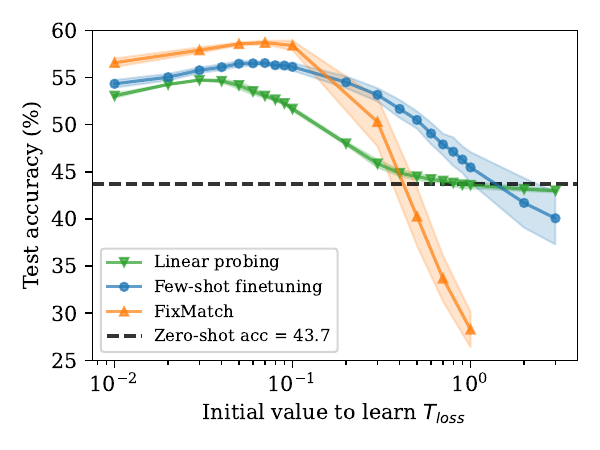}
\end{minipage}\hfill
\begin{minipage}{0.47\columnwidth}
    \captionsetup{type=figure}
\caption{\small
{\bf Impact of initial value to learn $T_\text{loss}$}.
We run experiments using the OpenCLIP ViT-B/32 model \cite{cherti2023reproducible} with 16-shot labeled data sampled with three random seeds from the semi-Aves dataset \cite{semi-aves}.
For SSFSL with FixMatch, we use $T_\text{conf}=0.01$ per \cref{fig:confidence_temp_fixmatch}.
Results validate that using a smaller loss temperature significantly improves learning with VLM, with a small standard deviation, regardless of linear probing, FSL, or SSFSL.
}
\label{fig:acc_w_varying_Tloss}
\end{minipage}
\end{figure}

\textbf{
Proper initialization of loss temperature improves VLM performance.}
\cref{fig:acc_w_varying_Tloss} compares the performance of different initial values for learning the loss temperature $T_\text{loss}$ when linear probing, few-shot finetuning, and semi-supervised finetuning a VLM. 
Results show that using a small temperature consistently yields significant accuracy gains, while learning $T_\text{loss}$ from 0.07 (same as CLIP's pretraining \cite{radford2021learning}) empirically performs the best for both few-shot finetuning and semi-supervised finetuning.

\textbf{Temperature is crucial for learning with a VLM.}
\cref{tab:ablate_ts_probing} extends \cref{tab:cls_stage1} by studying more temperature settings across different backbones, including the ResNet-50 and ViT-B/32 pretrained on ImageNet \cite{deng2009imagenet} and those from the VLM CLIP \cite{radford2021learning, cherti2023reproducible}.
Results confirm the importance of using the loss temperature for linear probing a VLM.
The same observations are made in \cref{tab:learn_temp_fsft_fixmatch}, which shows that using a small temperature significantly improves few-shot finetuning and semi-supervised finetuning with VLM.

{
\setlength{\tabcolsep}{1.6em}
\begin{table}[t]
\centering
\caption{\small
{\bf Temperature is crucial for linear probing a VLM.}
We run linear probing with a frozen backbone using 16-shot labeled data from semi-Aves \cite{semi-aves} and compare different pretrained ResNet-50 \cite{he2016deep} and ViT-B/32 models \cite{dosovitskiy2021image}.
For models pretrained on ImageNet with Cross-Entropy loss, we randomly initialize their classifier weights, while for contrastively pretrained VLMs, we use text embeddings of class names \cite{parashar2024neglected}. 
Results show that not using the loss temperature performs the best for ImageNet-pretrained backbones, while doing so leads to marginal improvements for CLIP models over the zero-shot accuracy, due to the weak supervision from ``flat softmax probabilities'' (\cref{fig:diagnosis}A).
In contrast, using a smaller loss temperature sharpens the logits distribution (\cref{fig:diagnosis}B), thus significantly improving the linear probing performance for VLMs.
{\bf Bold} and \underline{underlined} numbers mark the best and second best numeric metrics.
}
\vspace{-3mm}
\label{tab:ablate_ts_probing}
\scalebox{0.8}{
\begin{tabular}{lcccc}
\toprule
backbone & INet-RN50 & INet-ViT & CLIP-RN50 & CLIP-ViT \\

\midrule
zero-shot acc \cite{parashar2024neglected} & -- & -- & 36.8 & 43.8 \\
\midrule

$T_\text{loss} = 0.01$ & 30.1 & 26.0 & \underline{46.2} & \textbf{53.5} \\
$T_\text{loss} = 0.07$ & 38.8 & 27.4 & 45.4 & 52.0 \\
$T_\text{loss} = 0.1$ & \underline{39.4} & \underline{29.1} & 45.9 & 50.4 \\
No Temp ($T_\text{loss} = 1.0$) & \textbf{40.7} & \textbf{31.5} & 39.9 & 43.9 \\

\midrule

$T_\text{loss}$ learned from 0.01 & 31.3 & 26.4 & \textbf{46.4} & \underline{53.2} \\
$T_\text{loss}$ learned from 0.07 & 38.8 & 27.6 & 45.3 & \textbf{53.5} \\
$T_\text{loss}$ learned from 0.1 & \underline{39.4} & \underline{29.1} & 45.6 & 51.5 \\
$T_\text{loss}$ learned from 1.0 & \textbf{40.7} & \textbf{31.5} & 41.6 & 43.8 \\

\bottomrule
\end{tabular}}
\end{table}
}

\begin{table}[ht]
\centering
\caption{\small
{\bf Comparison of FSL and SSL performance across different pretrained backbones under various loss temperatures.}
We run experiments using 16-shot labeled data and unlabeled data from semi-Aves.
Results show that not using loss temperature performs the best for ImageNet-pretrained models, but results in significantly worse performance than linear probing for VLMs, due to weak supervision (\cref{fig:confidence_temp_fixmatch}).
In contrast, using a small loss temperature consistently improves both FSL and SSL performance across VLM backbones.
\textcolor{Green}{Superscripts} denote accuracy improvement relative to the corresponding linear probing accuracy of each model.
}
\vspace{-3mm}
\label{tab:learn_temp_fsft_fixmatch}
\setlength{\tabcolsep}{0.5mm}
\scalebox{0.75}{
\begin{tabular}{lllll  llll}

\toprule
 & \multicolumn{4}{c}{Few-Shot Learning (FSL)} & \multicolumn{4}{c}{Semi-Supervised Learning (SSL)} \\

\cmidrule(r){2-5} \cmidrule(r){6-9} 

\multicolumn{1}{c}{backbone} & INet-RN50 & INet-ViT & CLIP-RN50 & CLIP-ViT & INet-50 & INet-ViT & CLIP-RN50 & CLIP-ViT \\

\midrule
\multicolumn{1}{c}{linear probing} & 40.7 & 31.5 & 46.2 & 53.5 & 40.7 & 31.5 & 46.2 & 53.5 \\

\midrule
$T_\text{loss} = 0.01$ & \ \ 0.5$^{\textcolor{Red}{-40.2}}$ & \ \ 0.5$^{\textcolor{Red}{-31.0}}$ & 44.7$^{\textcolor{Red}{-1.5}}$ & 53.8$^{\textcolor{Green}{+0.3}}$ & \ \ 0.5$^{\textcolor{Red}{-40.2}}$ & \ \ 0.5$^{\textcolor{Red}{-31.0}}$ & 40.2$^{\textcolor{Red}{-6.0}}$ & 57.1$^{\textcolor{Green}{+3.6}}$ \\

$T_\text{loss} = 0.07$ & 41.4$^{\textcolor{Green}{+0.7}}$ & \underline{33.3}$^{\textcolor{Green}{+1.8}}$ & 50.8$^{\textcolor{Green}{+4.6}}$ & \underline{56.2}$^{\textcolor{Green}{+2.7}}$ & 39.6$^{\textcolor{Red}{-1.1}}$ & 33.1$^{\textcolor{Green}{+1.6}}$ & 47.8$^{\textcolor{Green}{+1.6}}$ & \underline{57.9}$^{\textcolor{Green}{+4.4}}$ \\

$T_\text{loss} = 0.1$ & 42.3$^{\textcolor{Green}{+1.6}}$ & 32.2$^{\textcolor{Green}{+0.7}}$ & \textbf{51.8}$^{\textcolor{Green}{+5.6}}$ & 55.9$^{\textcolor{Green}{+2.4}}$ & 40.4$^{\textcolor{Red}{-0.3}}$ & 31.3$^{\textcolor{Red}{-0.2}}$ & \textbf{48.7}$^{\textcolor{Green}{+2.5}}$ & 57.3$^{\textcolor{Green}{+3.8}}$ \\
 
No Temp ($T_\text{loss} = 1.0$)
 & \cellcolor{green!30}\textbf{43.4$^{\textcolor{Green}{+2.7}}$} 
 & \cellcolor{green!30}\textbf{33.7$^{\textcolor{Green}{+2.2}}$} 
 & \cellcolor{red!30}41.6$^{\textcolor{Red}{-4.6}}$ 
 & \cellcolor{red!30}42.1$^{\textcolor{Red}{-11.4}}$ 
 & \cellcolor{green!30}\underline{48.0}$^{\textcolor{Green}{+7.3}}$ 
 & \cellcolor{green!30}\textbf{37.1}$^{\textcolor{Green}{+5.6}}$ 
 & \cellcolor{red!30}22.5$^{\textcolor{Red}{-23.7}}$ 
 & \cellcolor{red!30}23.8$^{\textcolor{Red}{-29.7}}$ \\

\midrule
 
$T_\text{loss}$ learned from 0.01 & \ \ 0.5$^{\textcolor{Red}{-40.2}}$ & \ \ 0.6$^{\textcolor{Red}{-30.9}}$ & 44.7$^{\textcolor{Red}{-1.5}}$ & 54.1$^{\textcolor{Green}{+0.6}}$ & \ \ 0.5$^{\textcolor{Red}{-40.2}}$ & \ \ 0.6$^{\textcolor{Red}{-30.9}}$ & 40.5$^{\textcolor{Red}{-5.7}}$ & 57.0$^{\textcolor{Green}{+3.5}}$ \\

$T_\text{loss}$ learned from 0.07 & 41.7$^{\textcolor{Green}{+1.0}}$ & 32.6$^{\textcolor{Green}{+1.1}}$ 
 & \cellcolor{green!30}50.3$^{\textcolor{Green}{+4.1}}$ 
 & \cellcolor{green!30}\textbf{56.5}$^{\textcolor{Green}{+3.0}}$ & 39.9$^{\textcolor{Red}{-0.8}}$ & 32.6$^{\textcolor{Green}{+1.1}}$ 
 & \cellcolor{green!30}47.0$^{\textcolor{Green}{+0.8}}$ 
 & \cellcolor{green!30}\textbf{58.5}$^{\textcolor{Green}{+5.0}}$ \\
 
 $T_\text{loss}$ learned from 0.1 & 42.6$^{\textcolor{Green}{+1.9}}$ & 33.2$^{\textcolor{Green}{+1.7}}$ & \underline{51.7}$^{\textcolor{Green}{+5.5}}$ & \underline{56.2}$^{\textcolor{Green}{+2.7}}$ & 41.7$^{\textcolor{Green}{+1.0}}$ & 31.9$^{\textcolor{Green}{+0.4}}$ & \underline{48.3}$^{\textcolor{Green}{+2.1}}$ & \underline{57.9}$^{\textcolor{Green}{+4.4}}$ \\
 
$T_\text{loss}$ learned from 1.0 & \underline{43.3}$^{\textcolor{Green}{+2.6}}$ & 32.4$^{\textcolor{Green}{+0.9}}$ & 43.3$^{\textcolor{Red}{-2.9}}$ & 43.6$^{\textcolor{Red}{-9.9}}$ & \textbf{48.1}$^{\textcolor{Green}{+7.4}}$ & \underline{36.3}$^{\textcolor{Green}{+4.8}}$ & 26.5$^{\textcolor{Red}{-19.7}}$ & 26.1$^{\textcolor{Red}{-27.4}}$ \\

\bottomrule
\end{tabular}}
\vspace{-3mm}
\end{table}

\begin{table}[t]
\centering
\small
\caption{\small 
\textbf{SWIFT generalizes to different SSL methods.}
We experiment SWIFT with FixMatch \cite{sohn2020fixmatch} and DebiasPL \cite{wang2022debiased} for finetuning the OpenCLIP ViT-B/32 model \cite{cherti2023reproducible}.
Results show that SWIFT effectively improves both FixMatch and DebiasPL in each stage.
\textcolor{Green}{Superscripts} mark the incremental improvements relative to the previous row.
}
\vspace{-3mm}
\label{tab:swift_debiaspl}
\setlength{\tabcolsep}{4.0mm}
\scalebox{0.85}{
\begin{tabular}{lllll}
\toprule
\multirow{2}{*}{method} & \multirow{2}{*}{\makecell{training\\data}} & \multicolumn{3}{c}{mean acc. over five datasets} \\
\cmidrule{3-5}
&  &4-shot &8-shot &16-shot \\
\midrule

FS-FT \cite{liu2025few} \textsubscript{CVPR'25} & L & 61.0 & 65.8 & 69.1 \\
\rowcolor{col33}
SWAT \cite{liu2025few} \textsubscript{CVPR'25} & L+R &67.4 &71.0 &74.0 \\
\midrule

FixMatch\cite{sohn2020fixmatch} on VLM & L+U 
& 39.3 
& 49.9 
& 57.2 \\

\quad + stage-1: Classifier Init.  & L+U
& 56.3$^{\textcolor{Green}{+17.0}}$ 
& 59.8$^{\textcolor{Green}{+9.9}}$ 
& 62.2$^{\textcolor{Green}{+5.0}}$ \\

\quad \quad + stage-2: Semi-Sup. FT & L+U+R 
& 68.8$^{\textcolor{Green}{+11.1}}$ 
& 73.3$^{\textcolor{Green}{+7.6}}$ 
& 77.3$^{\textcolor{Green}{+6.1}}$ \\

\rowcolor{green!20}
\quad \quad \quad + stage-3: FS-FT (SWIFT)  & L+U+R 
& \textbf{71.5$^{\textcolor{Green}{+2.7}}$} 
& \textbf{76.3$^{\textcolor{Green}{+3.0}}$} 
& \textbf{79.7$^{\textcolor{Green}{+2.4}}$} \\

\midrule

DebiasPL\cite{wang2022debiased} on VLM & L+U 
&39.6  
&49.8  
&57.1  \\

\quad + stage-1: Classifier Init.  & L+U
& 56.3$^{\textcolor{Green}{+16.7}}$ 
& 59.9$^{\textcolor{Green}{+10.1}}$ 
& 62.1$^{\textcolor{Green}{+5.0}}$ \\

\quad \quad + stage-2: Semi-Sup. FT & L+U+R 
& 71.0$^{\textcolor{Green}{+10.7}}$ 
& 75.0$^{\textcolor{Green}{+7.4}}$ 
& 77.4$^{\textcolor{Green}{+4.2}}$ \\

\rowcolor{green!20}
\quad \quad \quad + stage-3: FS-FT (SWIFT)  & L+U+R 
& \textbf{73.1$^{\textcolor{Green}{+2.1}}$} 
& \textbf{77.4$^{\textcolor{Green}{+2.4}}$} 
& \textbf{79.9$^{\textcolor{Green}{+2.5}}$} \\

\bottomrule
\end{tabular}}
\vspace{-4mm}
\end{table}

\textbf{SWIFT generalizes to different SSL methods.}
\cref{tab:swift_debiaspl} shows that SWIFT can be integrated with stronger SSL methods, such as DebiaPL \cite{wang2022debiased}, which yields further improvements.

\textbf{SWIFT generalizes to DINOv2 backbone.}
\cref{tab:swift_dinov2_detail} shows the per-dataset performance when applying SWIFT to DINOv2 backbone. The consistent accuracy gains after each stage highlight SWIFT's generalization capability to different pretrained backbones.

\textbf{Impact of the number of retrieved data.}
\cref{tab:ablate_retrieved_images} shows that retrieving more data further improves SWIFT's performance, with consistently strong accuracy gains over prior state-of-the-art FSL method SWAT \cite{liu2025few}.

{
\setlength{\tabcolsep}{0.4em}
\begin{table*}[]
\centering
\caption{\small
{\bf Performance of SWIFT using DINOv2 backbone.} 
We experiment SWIFT with DINOv2 ViT-B/14 model \cite{oquab2024dinov} with 4-, 8-, and 16-shot labeled data. 
Results show that each component yields significant gains at each stage, validating that SWIFT effectively generalizes to different backbones. 
\textcolor{Green}{Superscripts} mark the incremental improvements relative to the previous row.
}
\vspace{-3mm}
\label{tab:swift_dinov2_detail}
\scalebox{0.8}{
\begin{tabular}{lllcccccl}
\toprule
shots & method & train data & semi-Aves & Aircraft & Cars & EuroSAT & DTD & mean acc. \\
\midrule
\midrule
\multirow{7}{*}{4} & FS-FT \cite{liu2025few} & L & 65.5 & 37.5 & 72.7 & 46.5 & 60.4 & 56.5 \\
\cmidrule{2-9}
 & FixMatch w/ DINOv2 & L+U & 61.6 & 31.8 & 66.9 & 39.1 & 51.8 & 50.2 \\
 & \quad + stage-1: cls init. & L+U & 65.8 & 37.6 & 73.3 & 46.5 & 60.3 & 56.7$^{\textcolor{Green}{+6.5}}$ \\
 & \quad \quad + stage-2: SS-FT & L+U+R & 83.5 & 80.1 & 87.5 & 61.0 & 69.5 & 76.3$^{\textcolor{Green}{+19.6}}$ \\
 & \quad \quad \quad + stage-3: FS-FT & L+U+R & 85.6 & 82.1 & 90.2 & 62.3 & 71.0 & 78.2$^{\textcolor{Green}{+1.9}}$ \\
\cmidrule{2-9}

 & \cellcolor{gray!15}Fully supervised & \cellcolor{gray!15}L+U & \cellcolor{gray!15}87.9 & \cellcolor{gray!15}82.7 & \cellcolor{gray!15}92.7 & \cellcolor{gray!15}98.9 & \cellcolor{gray!15}84.3 & \cellcolor{gray!15}89.3 \\

 & \cellcolor{gray!15}Fully supervised w/ RA & \cellcolor{gray!15}L+U+R & \cellcolor{gray!15}88.4 & \cellcolor{gray!15}88.5 & \cellcolor{gray!15}90.8 & \cellcolor{gray!15}98.8 & \cellcolor{gray!15}79.1 & \cellcolor{gray!15}89.1 \\
 \midrule
 \midrule
 
\multirow{7}{*}{8} & FS-FT \cite{liu2025few} & L & 74.4 & 52.4 & 83.3 & 74.6 & 70.9 & 71.1 \\
\cmidrule{2-9}
 & FixMatch w/ DINOv2 & L+U & 73.1 & 46.3 & 80.8 & 64.6 & 66.7 & 66.3 \\
 & \quad + stage-1: cls init. & L+U & 75.8 & 52.4 & 84.2 & 73.1 & 71.9 & 71.5$^{\textcolor{Green}{+5.2}}$ \\
 & \quad \quad + stage-2: SS-FT & L+U+R & 85.2 & 82.2 & 88.4 & 81.6 & 72.4 & 82.0$^{\textcolor{Green}{+10.5}}$ \\
 & \quad \quad \quad + stage-3: FS-FT & L+U+R & 86.9 & 84.3 & 91.7 & 84.4 & 74.6 & 84.4$^{\textcolor{Green}{+2.4}}$ \\
 \cmidrule{2-9}

 & \cellcolor{gray!15}Fully supervised & \cellcolor{gray!15}L+U & \cellcolor{gray!15}88.3 & \cellcolor{gray!15}83.1 & \cellcolor{gray!15}93.4 & \cellcolor{gray!15}99.0 & \cellcolor{gray!15}85.2 & \cellcolor{gray!15}89.8 \\

 & \cellcolor{gray!15}Fully supervised w/ RA & \cellcolor{gray!15}L+U+R & \cellcolor{gray!15}88.6 & \cellcolor{gray!15}88.9 & \cellcolor{gray!15}90.9 & \cellcolor{gray!15}98.9 & \cellcolor{gray!15}79.4 & \cellcolor{gray!15}89.3 \\
\midrule
\midrule

\multirow{7}{*}{16} & FS-FT \cite{liu2025few} & L & 79.7 & 66.3 & 90.1 & 85.2 & 76.5 & 79.6 \\
\cmidrule{2-9}
 & FixMatch w/ DINOv2 & L+U & 80.1 & 62.0 & 89.4 & 79.7 & 74.5 & 77.1 \\
 & \quad + stage-1: cls init. & L+U & 81.3 & 66.7 & 91.0 & 85.2 & 77.2 & 80.3$^{\textcolor{Green}{+3.1}}$ \\
 & \quad \quad + stage-2: SS-FT & L+U+R & 87.2 & 85.5 & 90.2 & 92.5 & 76.1 & 86.3$^{\textcolor{Green}{+6.0}}$ \\
 & \quad \quad \quad + stage-3: FS-FT & L+U+R & 88.2 & 86.5 & 93.7 & 93.0 & 77.7 & 87.8$^{\textcolor{Green}{+1.5}}$ \\
 \cmidrule{2-9}

 & \cellcolor{gray!15}Fully supervised & \cellcolor{gray!15}L+U & \cellcolor{gray!15}88.7 & \cellcolor{gray!15}84.6 & \cellcolor{gray!15}94.1 & \cellcolor{gray!15}99.0 & \cellcolor{gray!15}85.3 & \cellcolor{gray!15}90.3 \\
  
 & \cellcolor{gray!15}Fully supervised w/ RA & \cellcolor{gray!15}L+U+R & \cellcolor{gray!15}88.8 & \cellcolor{gray!15}89.3 & \cellcolor{gray!15}91.6 & \cellcolor{gray!15}98.8 & \cellcolor{gray!15}80.2 & \cellcolor{gray!15}89.7 \\
\midrule
\bottomrule
\end{tabular}
}
\end{table*}
}

{\setlength{\tabcolsep}{1.0em}
\begin{table}[ht]
    \centering
    \vspace{-3mm}
    \caption{\small
{\bf Impact of number of retrieved images.}
We compare the performance of SWIFT and the prior state-of-the-art FSL method, SWAT \cite{liu2025few}, using OpenCLIP on the 16-shot semi-Aves dataset. 
Note that both methods use the same retrieved data with varying quantities. Results show that using more retrieved images improves performance. Importantly, SWIFT consistently outperforms SWAT by $>$ 4\% accuracy, due to the successful exploitation of unlabeled data.
}
\vspace{-3mm}
\label{tab:ablate_retrieved_images}
\scalebox{0.9}{
\begin{tabular}{lllll}
\hline
retrieved img/cls & 100 & 300 & 500 (in main paper) & 1000 \\
\hline
SWAT \cite{liu2025few} (CVPR’25) & 61.8 & 63.2 & 63.1 & 63.6 \\
\textbf{SWIFT (ours)} & 66.2$^{\textcolor{Green}{+4.4}}$  & 68.0$^{\textcolor{Green}{+4.8}}$ 
& 68.7$^{\textcolor{Green}{+5.6}}$ 
& 69.7$^{\textcolor{Green}{+6.1}}$ \\
\hline
\end{tabular}}
\end{table}}

\section{More Detailed Results}
\label{sec:details}

We present per-dataset benchmarking results in \cref{tab:compare_sota_detail}, highlighting that SWIFT achieves state-of-the-art SSFSL performance.
Additionally, we provide a more detailed ablation study of SWIFT in \cref{tab:ablate_swift_detail}, demonstrating the effectiveness of each component in our design.

\section{Pseudo-code for Temperatures}
\label{sec:pseudocode}

\begin{figure*}[t]
\centering
\rule{\textwidth}{0.8pt}

\includegraphics[width=0.99 \textwidth, clip=true, trim = 4mm 0mm 6mm 0mm]{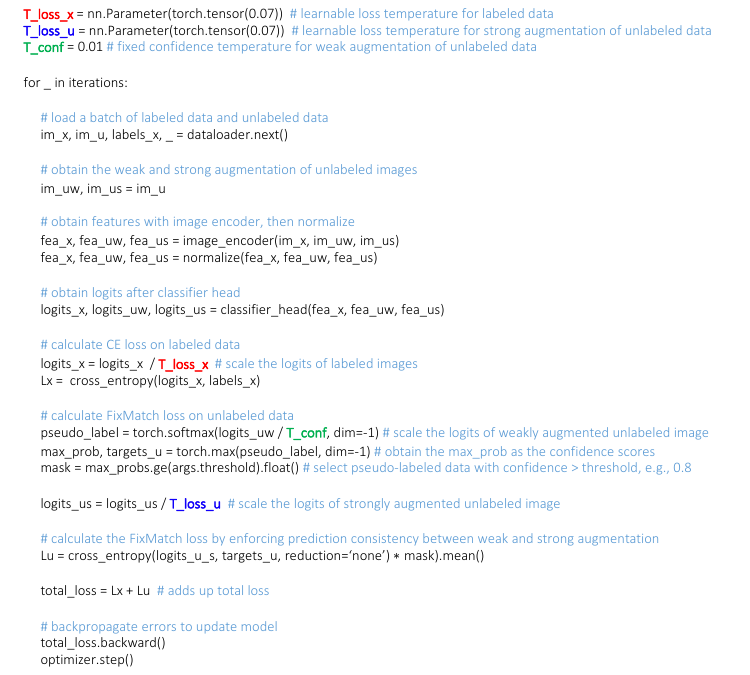} 
\rule{\textwidth}{0.8pt}

\vspace{-2mm}
\caption{\small
{\bf Example PyTorch-style pseudo-code for Temperatures in stage-2 semi-supervised finetuning of SWIFT.}
We illustrate our simple temperatures with FixMatch.
Specifically, we apply (1) a learnable \textcolor{Red}{loss temperature $T_{\text{loss}\_x}$} initialized to 0.07 to calculate the cross-entropy loss of labeled data; 
(2) a learnable \textcolor{blue}{loss temperature $T_{\text{loss}\_u}$} initialized to 0.07 to scale the logits of strongly augmented unlabeled images for calculating the consistency loss;
(3) a fixed \textcolor{Green}{confidence temperature $T_\text{conf}$} set to 0.01 to scale the confidence scores of weakly augmented unlabeled images to enable utilization of pseudo-labels. 
Importantly, our temperatures add negligible compute costs to the existing FixMatch \cite{sohn2020fixmatch} training recipe, yet contribute significant performance improvements for FSL and SSFSL with VLMs (cf. Table \ref{tab:compare_sota}).
}
\label{fig:pseudocode}
\end{figure*}

Fig.~\ref{fig:pseudocode} presents the pseudo-code of our temperatures with FixMatch \cite{sohn2020fixmatch}
We include the implementation of temperatures with DebiasPL \cite{wang2022debiased} in our code.

\section{Code and Instructions}
\label{sec:Demo-code}

We release open-source Python code 
at \url{https://github.com/tian1327/SWIFT} and provide usage instructions below.


{\bf Dependencies}.
Running our code requires some packages, including
\emph{clip}, \emph{open\_clip\_torch}, \emph{img2dataset}, \emph{torchvision}, and \emph{PyTorch}.
We suggest assigning $>$50GB storage space and $>$40GB GPU RAM to reproduce our experiments.

{\bf License}.
Code is released under the MIT License to foster future research.

{\bf Instructions.}
Detailed instructions for running our code are provided in the following markdown files.

\begin{itemize}
\item
\begin{verbatim}ENV.md\end{verbatim}
Create a conda environment and install the required packages.

\item
\begin{verbatim}DATASETS.md\end{verbatim}
Detailed steps for setting up the benchmarking datasets are provided.

\item
\begin{verbatim}RETRIEVAL.md\end{verbatim}
We provide detailed instructions, following \cite{liu2025few}, for using string-matching~\cite{parashar2024neglected} to retrieve task-relevant open images from OpenCLIP's pretraining dataset LAION-400M~\cite{laion400m, laion5b}.

\item
\begin{verbatim}README.md\end{verbatim}
We provide instructions to run the code for few-shot finetuning (FSFT) and SWIFT. In addition, we provide guidelines on how to reproduce the SSL baselines, including FixMatch~\cite{sohn2020fixmatch}, DebiasPL~\cite{wang2022debiased}, and FineSSL \cite{gan2024erasing}.

\end{itemize}

{
\setlength{\tabcolsep}{0.60em}
\begin{table}[ht]
\centering
\caption{\small
{\bf Detailed benchmarking results for each dataset.} 
We compare our SWIFT with recent FSL and SSFSL methods on five fine-grained datasets, with 4-, 8-, and 16-shot labeled data.
SWIFT consistently yields SOTA SSFSL performance.
}
\vspace{-3mm}
\label{tab:compare_sota_detail}
\scalebox{0.63}{
\begin{tabular}{ccllcccccc}
\toprule
shot & & method & strategy & semi-Aves & Aircraft & Cars & EuroSAT & DTD & mean acc. \\
\midrule
\midrule

\multirow{21}{*}{4} & \multirow{10}{*}{\rotatebox[origin=c]{90}{FSL}}  & CoOp \cite{zhou2022learning} \textsubscript{IJCV'22} & \textcolor{Gray}{prompt tuning} & 38.1 & 20.6 & 62.7 & 68.6 & 53.9 & 48.8 \\
 &  & PLOT \cite{chen2022plot} \textsubscript{ICLR'23} & \textcolor{Gray}{prompt tuning} & 37.2 & 22.4 & 63.4 & 72.4 & 56.0 & 50.3 \\
 &  & Linear Probing \cite{radford2021learning} \textsubscript{ICML'21} & \textcolor{Gray}{adapter learning} & 47.0 & 24.6 & 80.7 & 68.9 & 63.9 & 57.0 \\
 &  & CLIP-Adapter \cite{clipadapter} \textsubscript{IJCV'23} & \textcolor{Gray}{adapter learning} & 39.2 & 23.0 & 61.0 & 72.5 & 47.2 & 48.6 \\
 &  & Tip-Adapter(f) \cite{tipadapter} \textsubscript{ECCV'22} & \textcolor{Gray}{adapter learning} & 42.4 & 21.9 & 61.1 & 66.8 & 58.0 & 50.0 \\
 &  & TaskRes(e) \cite{yu2023task} \textsubscript{ECCV'22} & \textcolor{Gray}{adapter learning} & 43.2 & 25.9 & 64.7 & 73.0 & 58.4 & 53.0 \\
 &  & CMLP \cite{lin2023multi} \textsubscript{CVPR'23} & \textcolor{Gray}{adapter learning} & 29.1 & 25.1 & 80.7 & 74.8 & 62.2 & 54.4 \\
 &  & CLAP \cite{clap24} \textsubscript{CVPR'24} & \textcolor{Gray}{adapter learning} & 34.0 & 28.0 & 84.9 & 74.7 & 63.0 & 56.9 \\
 &  & Few-Shot FT \cite{liu2025few} \textsubscript{CVPR'25} & \textcolor{Gray}{finetune} & 48.0 & 28.8 & 82.5 & 81.8 & 66.7 & 61.6 \\
 &  & \cellcolor{col33}SWAT \cite{liu2025few} \textsubscript{CVPR'25} & \cellcolor{col33}\textcolor{Gray}{finetune w/ RA} & \cellcolor{col33}58.5 & \cellcolor{col33}55.7 & \cellcolor{col33}81.1 & \cellcolor{col33}83.2 & \cellcolor{col33}58.3 & \cellcolor{col33}67.4 \\
 
 \cmidrule(r){2-10}
 
 & \multirow{8}{*}{\rotatebox[origin=c]{90}{SSFSL}}  & FixMatch (IN-RN50) \cite{sohn2020fixmatch} \textsubscript{NeurIPS'22}  
 & \textcolor{Gray}{finetune} & 25.3 & 16.1 & 17.2 & 50.3 & 44.8 & 30.7 \\

 &  & FixMatch (VLM-ViT) & \textcolor{Gray}{finetune} &  19.0 &  18.6 &  56.0 &  55.5 &  47.2 &  39.3 \\

& & \cellcolor{green!20}FixMatch (ours) & \cellcolor{green!20}\textcolor{Gray}{finetune} & \cellcolor{green!20}45.6 & \cellcolor{green!20}20.5 & \cellcolor{green!20}78.4 & \cellcolor{green!20}80.2 & \cellcolor{green!20}63.6 & \cellcolor{green!20}57.7 \\

\cmidrule{3-10}

 &  & DebiasPL (IN-RN50) \cite{wang2022debiased}  \textsubscript{CVPR'22} 
 & \textcolor{Gray}{finetune} & 22.9 & 15.3 & 20.4 & 64.5 & 49.2 & 34.5 \\

  &  &  DebiasPL (VLM-ViT) &  \textcolor{Gray}{finetune} &  19.2 &  18.6 &  56.3 &  56.7 &  47.2 &  39.6 \\

 &  & \cellcolor{green!20}DebiasPL (ours) & \cellcolor{green!20}\textcolor{Gray}{finetune} & \cellcolor{green!20}47.7 & \cellcolor{green!20}24.2 & \cellcolor{green!20}83.7 & \cellcolor{green!20}80.8 & \cellcolor{green!20}65.0 & \cellcolor{green!20}60.3 \\
\cmidrule{3-10}

 &  & FineSSL (VLM-ViT) \cite{gan2024erasing} \textsubscript{ICML'24} 
 & \textcolor{Gray}{{prompt tuning}} & 28.0 & 21.8 & 80.9 & 94.9 & 62.7 & 57.6 \\

 &  & \cellcolor{green!20}SWIFT (ours)& \cellcolor{green!20}\textcolor{Gray}{finetune w/ RA} & \cellcolor{green!20}65.8 & \cellcolor{green!20}64.4 & \cellcolor{green!20}88.9 & \cellcolor{green!20}75.0 & \cellcolor{green!20}63.5 & \cellcolor{green!20}71.5 \\

  \cmidrule(r){2-10}
  
 & \multirow{2}{*}{\rotatebox[origin=c]{90}{Ref.}}  & \cellcolor{gray!15}Fully supervised & \cellcolor{gray!15}\textcolor{Gray}{finetune} & \cellcolor{gray!15}65.3 & \cellcolor{gray!15}44.8 & \cellcolor{gray!15}87.4 & \cellcolor{gray!15}99.0 & \cellcolor{gray!15}77.7 & \cellcolor{gray!15}74.8 \\
 &  & \cellcolor{gray!15}Fully supervised w/ RA & \cellcolor{gray!15}\textcolor{Gray}{finetune w/ RA} & \cellcolor{gray!15}66.8 & \cellcolor{gray!15}58.8 & \cellcolor{gray!15}82.7 & \cellcolor{gray!15}98.9 & \cellcolor{gray!15}72.9 & \cellcolor{gray!15}76.0 \\

\midrule
\midrule

\multirow{21}{*}{8} & \multirow{10}{*}{\rotatebox[origin=c]{90}{FSL}} & CoOp \cite{zhou2022learning} \textsubscript{IJCV'22} & \textcolor{Gray}{prompt tuning} & 42.0 & 26.6 & 67.6 & 77.1 & 59.7 & 54.6 \\
 &  & PLOT \cite{chen2022plot} \textsubscript{ICLR'23} & \textcolor{Gray}{prompt tuning} & 41.4 & 26.2 & 67.0 & 78.2 & 61.7 & 54.9 \\
 &  & Linear Probing \cite{radford2021learning} \textsubscript{ICML'21} & \textcolor{Gray}{adapter learning} & 50.7 & 28.7 & 82.3 & 76.5 & 67.7 & 61.2 \\
 &  & CLIP-Adapter \cite{clipadapter} \textsubscript{IJCV'23} & \textcolor{Gray}{adapter learning} & 41.2 & 27.9 & 66.8 & 78.5 & 61.4 & 55.2 \\
 &  & Tip-Adapter \cite{tipadapter} \textsubscript{ECCV'22} & \textcolor{Gray}{adapter learning} & 46.2 & 23.8 & 64.4 & 70.3 & 59.8 & 52.9 \\
 &  & TaskRes(e) \cite{yu2023task} \textsubscript{ECCV'22} & \textcolor{Gray}{adapter learning} & 47.1 & 30.9 & 69.7 & 78.8 & 63.5 & 58.0 \\
 &  & CMLP \cite{lin2023multi} \textsubscript{CVPR'23} & \textcolor{Gray}{adapter learning} & 38.8 & 27.9 & 82.7 & 80.6 & 67.2 & 59.4 \\
 &  & CLAP \cite{clap24} \textsubscript{CVPR'24} & \textcolor{Gray}{adapter learning} & 42.9 & 33.6 & 86.1 & 77.4 & 66.4 & 61.3 \\
 &  & Few-Shot FT \cite{liu2025few} \textsubscript{CVPR'25} & \textcolor{Gray}{finetune} & 52.3 & 35.4 & 85.3 & 89.4 & 70.6 & 66.6 \\
 &  & \cellcolor{col33}SWAT \cite{liu2025few} \textsubscript{CVPR'25} & \cellcolor{col33}\textcolor{Gray}{finetune w/ RA} & \cellcolor{col33}60.8 & \cellcolor{col33}59.1 & \cellcolor{col33}83.5 & \cellcolor{col33}89.2 & \cellcolor{col33}62.6 & \cellcolor{col33}71.0 \\

 \cmidrule(r){2-10}

 & \multirow{8}{*}{\rotatebox[origin=c]{90}{SSFSL}}  & FixMatch (IN-RN50) \cite{sohn2020fixmatch} \textsubscript{NeurIPS'22}  
 & \textcolor{Gray}{finetune} & 37.6 & 25.4 & 31.0 & 76.8 & 55.9 & 45.3 \\

 &  & FixMatch (VLM-ViT) & \textcolor{Gray}{finetune} & 26.1 & 23.7 & 67.2 & 72.8 & 59.5 & 49.9 \\

 &  & \cellcolor{green!20}FixMatch (ours) & \cellcolor{green!20}\textcolor{Gray}{finetune} & \cellcolor{green!20}53.0 & \cellcolor{green!20}31.2 & \cellcolor{green!20}85.6 & \cellcolor{green!20}90.1 & \cellcolor{green!20}68.5 & \cellcolor{green!20}65.7 \\

 \cmidrule{3-10}
 
 &  & DebiasPL (IN-RN50) \cite{wang2022debiased}  \textsubscript{CVPR'22} 
 & \textcolor{Gray}{finetune} & 35.6 & 28.3 & 40.8 & 82.5 & 58.0 & 49.0 \\

 &  & DebiasPL (VLM-ViT) & \textcolor{Gray}{finetune} & 26.1 & 23.5 & 67.3 & 72.7 & 59.3 & 49.8 \\
 
 &  & \cellcolor{green!20}DebiasPL (ours) & \cellcolor{green!20}\textcolor{Gray}{finetune} & \cellcolor{green!20}53.3 & \cellcolor{green!20}35.0 & \cellcolor{green!20}88.1 & \cellcolor{green!20}91.2 & \cellcolor{green!20}70.4 & \cellcolor{green!20}67.6 \\
 
  \cmidrule(r){3-10}
 
 &  & FineSSL (VLM-ViT) \cite{gan2024erasing} \textsubscript{ICML'24} & \textcolor{Gray}{prompt tuning} & 41.3 & 31.4 & 86.1 & 96.4 & 67.7 & 64.6 \\

 &  & \cellcolor{green!20}SWIFT (ours)& \cellcolor{green!20}\textcolor{Gray}{finetune w/ RA} & \cellcolor{green!20}67.7 & \cellcolor{green!20}67.7 & \cellcolor{green!20}90.6 & \cellcolor{green!20}87.1 & \cellcolor{green!20}68.5 & \cellcolor{green!20}76.3 \\

  \cmidrule(r){2-10}
 & \multirow{2}{*}{\rotatebox[origin=c]{90}{Ref.}}  & \cellcolor{gray!15}Fully supervised & \cellcolor{gray!15}\textcolor{Gray}{finetune} & \cellcolor{gray!15}65.6 & \cellcolor{gray!15}46.0 & \cellcolor{gray!15}88.2 & \cellcolor{gray!15}99.2 & \cellcolor{gray!15}78.0 & \cellcolor{gray!15}75.4 \\
 &  & \cellcolor{gray!15}Fully supervised w/ RA & \cellcolor{gray!15}\textcolor{Gray}{finetune w/ RA} & \cellcolor{gray!15}66.9 & \cellcolor{gray!15}60.4 & \cellcolor{gray!15}83.3 & \cellcolor{gray!15}99.0 & 7\cellcolor{gray!15}4.6 & \cellcolor{gray!15}76.8 \\

\midrule
\midrule

\multirow{21}{*}{16} & \multirow{10}{*}{\rotatebox[origin=c]{90}{FSL}} & CoOp \cite{zhou2022learning} \textsubscript{IJCV'22} & \textcolor{Gray}{prompt tuning} & 46.1 & 31.4 & 73.6 & 83.7 & 62.5 & 59.5 \\
 &  & PLOT \cite{chen2022plot} \textsubscript{ICLR'23} & \textcolor{Gray}{prompt tuning} & 44.4 & 31.5 & 72.8 & 82.2 & 65.6 & 59.3 \\
 &  & Linear Probing \cite{radford2021learning} \textsubscript{ICML'21} & \textcolor{Gray}{adapter learning} & 53.5 & 31.7 & 84.3 & 82.0 & 71.8 & 64.7 \\
 &  & CLIP-Adapter \cite{clipadapter} \textsubscript{IJCV'23} & \textcolor{Gray}{adapter learning} & 43.6 & 34.2 & 73.5 & 83.2 & 65.7 & 60.0 \\
 &  & Tip-Adapter \cite{tipadapter} \textsubscript{ECCV'22} & \textcolor{Gray}{adapter learning} & 50.1 & 29.3 & 69.6 & 76.6 & 64.6 & 58.0 \\
 &  & TaskRes(e) \cite{yu2023task} \textsubscript{ECCV'22} & \textcolor{Gray}{adapter learning} & 48.5 & 36.5 & 75.4 & 83.7 & 65.9 & 62.0 \\
 &  & CMLP \cite{lin2023multi} \textsubscript{CVPR'23} & \textcolor{Gray}{adapter learning} & 46.8 & 32.4 & 84.7 & 85.2 & 71.9 & 64.2 \\
 &  & CLAP \cite{clap24} \textsubscript{CVPR'24} & \textcolor{Gray}{adapter learning} & 49.2 & 39.1 & 87.8 & 81.7 & 69.9 & 65.5 \\
 &  & Few-Shot FT \cite{liu2025few} \textsubscript{CVPR'25} & \textcolor{Gray}{finetune} & 56.5 & 42.7 & 87.8 & 94.3 & 73.4 & 70.9 \\
 &  & \cellcolor{col33}SWAT \cite{liu2025few} \textsubscript{CVPR'25} & \cellcolor{col33}\textcolor{Gray}{finetune w/ RA} & \cellcolor{col33}63.1 & \cellcolor{col33}62.4 & \cellcolor{col33}85.4 & \cellcolor{col33}92.6 & \cellcolor{col33}66.3 & \cellcolor{col33}74.0 \\

\cmidrule(r){2-10}
 & \multirow{8}{*}{\rotatebox[origin=c]{90}{SSFSL}}  & FixMatch (IN-RN50) \cite{sohn2020fixmatch} \textsubscript{NeurIPS'22}  
 & \textcolor{Gray}{finetune} & 48.0 & 45.4 & 59.5 & 84.2 & 63.6 & 60.1 \\

 &  & FixMatch (VLM-ViT) & \textcolor{Gray}{finetune} & 31.8 & 28.4 & 71.0 & 87.9 & 67.0 & 57.2 \\

  &  & \cellcolor{green!20}FixMatch (ours) & \cellcolor{green!20}\textcolor{Gray}{finetune} & \cellcolor{green!20}58.7 & \cellcolor{green!20}40.3 & \cellcolor{green!20}90.5 & \cellcolor{green!20}92.7 & \cellcolor{green!20}73.8 & \cellcolor{green!20}71.2 \\

\cmidrule(r){3-10}
 
 &  & DebiasPL (IN-RN50) \cite{wang2022debiased}  \textsubscript{CVPR'22} 
 & \textcolor{Gray}{finetune} & 47.1 & 45.6 & 66.9 & 88.7 & 65.4 & 62.7 \\

 &  & DebiasPL (VLM-ViT) & \textcolor{Gray}{finetune} & 31.5 & 28.1 & 71.3 & 87.8 & 66.9 & 57.1 \\
 
 &  & \cellcolor{green!20}DebiasPL (ours) & \cellcolor{green!20}\textcolor{Gray}{finetune} & \cellcolor{green!20}58.7 & \cellcolor{green!20}47.7 & \cellcolor{green!20}91.2 & \cellcolor{green!20}94.0 & \cellcolor{green!20}74.2 & \cellcolor{green!20}73.2 \\

\cmidrule(r){3-10}

 &  & FineSSL (VLM-ViT) \cite{gan2024erasing} \textsubscript{ICML'24} & \textcolor{Gray}{prompt tuning} & 48.9 & 39.4 & 88.2 & 96.5 & 71.6 & 68.9 \\

 &  & \cellcolor{green!20}SWIFT (ours)& \cellcolor{green!20}\textcolor{Gray}{finetune w/ RA} & \cellcolor{green!20}68.7 & \cellcolor{green!20}71.0 & \cellcolor{green!20}92.7 & \cellcolor{green!20}94.7 & \cellcolor{green!20}71.4 & \cellcolor{green!20}79.7 \\

  \cmidrule(r){2-10}
 & \multirow{2}{*}{\rotatebox[origin=c]{90}{Ref.}}  & \cellcolor{gray!15}Fully supervised & \cellcolor{gray!15}\textcolor{Gray}{finetune} & \cellcolor{gray!15}66.3 & \cellcolor{gray!15}47.4 & \cellcolor{gray!15}89.0 & \cellcolor{gray!15}99.0 & \cellcolor{gray!15}78.3 & \cellcolor{gray!15}76.0 \\
 &  & \cellcolor{gray!15}Fully supervised w/ RA & \cellcolor{gray!15}\textcolor{Gray}{finetune w/ RA} & \cellcolor{gray!15}67.7 & \cellcolor{gray!15}60.1 & \cellcolor{gray!15}84.4 & \cellcolor{gray!15}98.9 & \cellcolor{gray!15}74.9 & \cellcolor{gray!15}77.2 \\
\midrule
\bottomrule
\end{tabular}
}
\end{table}
}

{
\setlength{\tabcolsep}{0.2em}
\begin{table*}[t]
\centering
\vspace{+4mm}
\caption{\small
{\bf Detailed ablation study result for each dataset.} 
We experiment with the OpenCLIP ViT-B/32 model \cite{cherti2023reproducible} with 4-, 8-, and 16-shot labeled data, and compare with the results of directly applying FixMatch \cite{sohn2020fixmatch} on VLM.
Results validate that each component of SWIFT brings significant performance gains.
{\bf Bold} and \underline{underlined} numbers mark the best and second best results.
}
\vspace{-2mm}
\label{tab:ablate_swift_detail}
\scalebox{0.78}{
\begin{tabular}{clccccccccccl}
\toprule
\multirow{2}{*}{shots} & \multirow{2}{*}{method} & stage 1 & \multicolumn{3}{c}{stage 2} & stage 3 & \multicolumn{6}{c}{datasets} \\

\cmidrule(r){3-3} \cmidrule(r){4-6} \cmidrule(r){7-7} \cmidrule(r){8-13}

& & \makecell{cls\\init.} & $T_\text{loss}$ & \makecell{retrieved\\data} & $T_\text{conf}$ & \makecell{few-shot\\FT} & {semi-Aves} & {Aircraft} & {Cars} & {EuroSAT} & {DTD} & {mean acc.} \\
 \midrule
\rowcolor{gray!15}\multirow{6}{*}{4} & FixMatch &  &  &  &  &  & 19.0 & 18.6 & 56.0 & 55.5 & 47.2 & 39.3 \\
 &  & \checkmark &  &  &  &  & 37.5 & 22.8 & 73.4 & 81.8 & 66.1 & 56.3$^{\textcolor{Green}{+17.0}}$ \\
 &  & \checkmark & \checkmark &  &  &  & 47.0 & 28.8 & 80.9 & 83.3 & 67.7 & 61.5$^{\textcolor{Green}{+5.2}}$ \\
 &  & \checkmark & \checkmark & \checkmark &  &  & 55.2 & 50.1 & 77.1 & 83.9 & 57.2 & 64.7$^{\textcolor{Green}{+3.2}}$ \\
 &  & \checkmark & \checkmark & \checkmark & \checkmark &  & 62.7 & 63.1 & 84.3 & 72.0 & 61.9 & \underline{68.8}$^{\textcolor{Green}{+4.1}}$ \\
 & \textbf{SWIFT} & \checkmark & \checkmark & \checkmark & \checkmark & \checkmark & 65.8 & 64.4 & 88.9 & 75.0 & 63.5 & \textbf{71.5}$^{\textcolor{Green}{+2.7}}$ \\
  
\midrule
  
\rowcolor{gray!15}\multirow{6}{*}{8} & FixMatch &  &  &  &  &  & 26.1 & 23.7 & 67.2 & 72.8 & 59.5 & 49.9 \\
 &  & \checkmark &  &  &  &  & 39.8 & 27.2 & 76.1 & 89.0 & 66.9 & 59.8$^{\textcolor{Green}{+9.9}}$ \\
 &  & \checkmark & \checkmark &  &  &  & 52.0 & 34.6 & 84.3 & 90.5 & 72.7 & 66.8$^{\textcolor{Green}{+7.0}}$ \\
 &  & \checkmark & \checkmark & \checkmark &  &  & 57.4 & 53.3 & 78.5 & 88.0 & 59.6 & 67.4$^{\textcolor{Green}{+0.5}}$ \\
 &  & \checkmark & \checkmark & \checkmark & \checkmark &  & 65.5 & 66.8 & 85.5 & 82.9 & 65.6 & \underline{73.3}$^{\textcolor{Green}{+5.9}}$ \\
 &\textbf{SWIFT} & \checkmark & \checkmark & \checkmark & \checkmark & \checkmark & 67.7 & 67.7 & 90.6 & 87.1 & 68.5 & \textbf{76.3}$^{\textcolor{Green}{+3.1}}$ \\
  
  \midrule
\rowcolor{gray!15}\multirow{6}{*}{16} & FixMatch &  &  &  &  &  & 31.8 & 28.4 & 71.0 & 87.9 & 67.0 & 57.2 \\
 &  & \checkmark &  &  &  &  & 42.0 & 30.0  & 78.1 & 93.4 & 67.7 & 62.2$^{\textcolor{Green}{+5.0}}$ \\
 &  & \checkmark & \checkmark &  &  &  & 56.4 & 40.1 & 86.4 & 93.6 & 73.6 & 70.0$^{\textcolor{Green}{+7.8}}$ \\
 &  & \checkmark & \checkmark & \checkmark &  &  & 60.0 & 55.6 & 80.2 & 92.1 & 63.5 & 70.3$^{\textcolor{Green}{+0.3}}$ \\
 &  & \checkmark & \checkmark & \checkmark & \checkmark &  & 67.0 & 69.7 & 87.7 & 93.9 & 68.1 & \underline{77.3}$^{\textcolor{Green}{+7.0}}$ \\
 
 & \textbf{SWIFT} & \checkmark & \checkmark & \checkmark & \checkmark & \checkmark & 68.7 & 71.0 & 92.7 & 94.7 & 71.4 & \textbf{79.7}$^{\textcolor{Green}{+2.4}}$ \\
\bottomrule
\end{tabular}
}
\end{table*}
}

\end{document}